\pdfoutput=1
\documentclass[journal]{IEEEtran}
\usepackage{graphicx}
\usepackage{amsmath,bm}
\usepackage{epstopdf}
\usepackage{multirow}
\usepackage{footmisc}

\ifCLASSOPTIONcompsoc
\usepackage[caption=false, font=normalsize, labelfont=sf, textfont=sf]{subfig}
\else
\usepackage[caption=false, font=footnotesize]{subfig}
\fi

\ifCLASSOPTIONcompsoc
\usepackage[nocompress]{cite}
\else
\usepackage{cite}
\fi

\ifCLASSINFOpdf
\else
\fi

\begin{document}
\title{PISEP${^2}$: Pseudo Image Sequence Evolution based ${3}$D Pose Prediction}

\author{Xiaoli~Liu,
        Jianqin~Yin,
        Huaping~Liu,
        and~Yilong~Yin % <-this % stops a space  ~\IEEEmembership{Member,~IEEE,}
\thanks{Xiaoli Liu and Jianqin Yin are with the Automation School of Beijing University of Posts
and Telecommunications, Beijing 100876, China (Liuxiaoli134@bupt.edu.cn; jqyin@bupt.edu.cn). (Corresponding author: Jianqin Yin)}% <-this % stops a space
\thanks{Huaping Liu is with the Department of Computer Science and Technology of Tsinghua University, Beijing 100084, China.}%
\thanks{Yilong Yin is with the school of Computer Science and Technology of
Shandong University, Jinan 250022, China.}}% <-this % stops a space


\markboth{Journal of \LaTeX\ Class Files,~Vol.~14, No.~8, August~2015}%
{Shell \MakeLowercase{\textit{et al.}}: Bare Demo of IEEEtran.cls for IEEE Journals}



\maketitle

\begin{abstract}
Pose prediction is to predict future poses given a window of previous poses. In this paper, we propose a new problem that predicts poses using 3D joint coordinate sequences. Different from the traditional pose prediction based on Mocap frames, this problem is convenient to use in real applications due to its simple sensors to capture data. We also present a new framework, PISEP${^2}$ (Pseudo Image Sequence Evolution based ${3}$D Pose Prediction), to address this new problem. Specifically, a skeletal representation is proposed by transforming the joint coordinate sequence into an image sequence, which can model the different correlations of different joints. With this image based skeletal representation, we model the pose prediction as the evolution of image sequence. Moreover, a novel inference network is proposed to predict all future poses in one step by decoupling the decoders in a non-recursive manner. Compared with the recursive sequence to sequence model, we can improve the computational efficiency and avoid error accumulation significantly. Extensive experiments are carried out on two benchmark datasets (e.g. G${3}$D and FNTU). The proposed method achieves the state-of-the-art performance on both datasets, which demonstrates the effectiveness of our proposed method.
\end{abstract}


% \fi
\IEEEpeerreviewmaketitle

\section{Introduction}
\begin{figure}[!t]
\centering
\includegraphics[width=\linewidth]{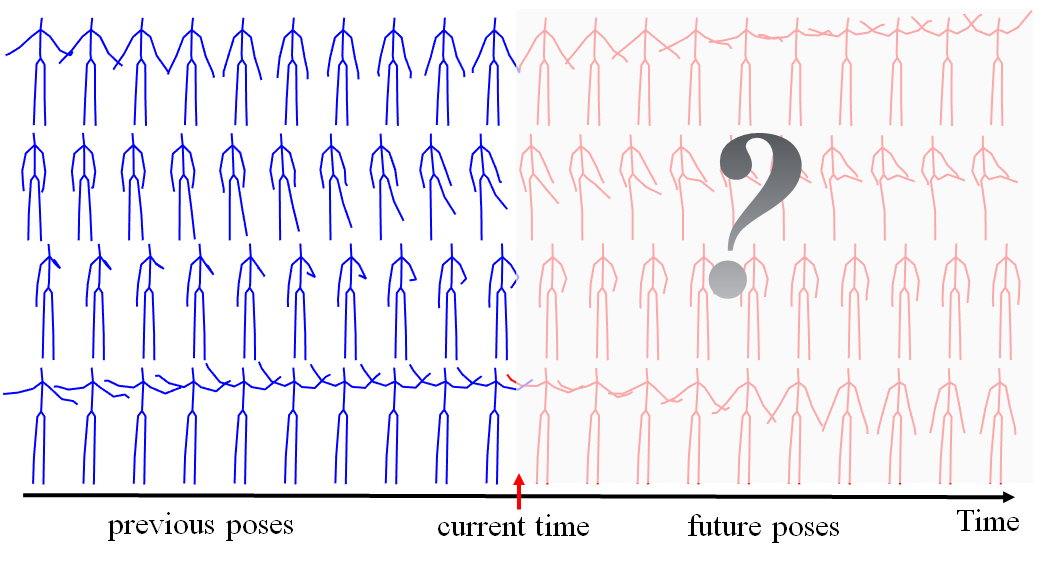}
\caption{pose prediction.}
\label{pp}
\end{figure}

\IEEEPARstart{P}{ose} prediction is widely applied to human-computer collaboration, family service robots, intelligent security and so on \cite{APsurvey}. It is important to predict future dynamics before it happens, which can provide more time for the robot to react and prepare ahead. With the development of the low-cost depth sensor (such as Kinect) and the 3D human pose estimation technique \cite{HanKinect,Tome3DPE}, we can easily and effectively acquire 3D skeletal data of humans. Moreover, joints position representation is effective, which is closely matches the visual dissimilarity in Euclidean space \cite{HoldenDM}. Therefore, we present to predict poses using a 3D joint coordinate sequence. As is shown in Figure \ref{pp}, the blue poses are the previous poses, and the red poses are the future poses. Our goal is to predict future joint coordinate sequence given a window of previous joint coordinate sequence.

{\bf From the perspective of the problem}, the most similar existing works are mocap based pose prediction and video prediction, which intrinsically belong to the sequence to sequence modeling. However, the input and output are different from ours. (1) {\bf Mocap based pose prediction} \cite{MartinezOMP,tprnn,rnmhy,srnnap}: on one hand, the input and output of these works are mocap frames, and the human pose is represented as a mocap vector parameterized by the exponential map which is easy to predict to a great extent \cite{ppruem,TaylorMM,h36m}; on the other hand, the acquisition of mocap data is difficult and expensive, and it needs lots of preprocessing to visualize its performance \cite{TaylorMM,MartinezOMP},which is inefficient and not intuitive. Therefore, this motivates us to predict human poses using joint coordinate sequence, which can be acquired cheaply, easily and efficiently. Moreover, due to the various distances or views of the placement of the camera, the problem of pose prediction with a joint coordinate sequence is very challenging. For example, the ${x}$ or ${y}$-axis joint coordinate value of the nearby will be larger than the distant. It will lead to the different physical structure characteristics of the human body. Therefore, mocap based pose prediction and joint coordinate sequence based pose prediction are two different problems, which are not comparable. (${2}$) {\bf Video prediction}: the input and output of video prediction is the image frame, which is different from ours. Specially, the 3D skeletal data has a sparse data structure, which is very different from the dense video data. This may lead to a huge gap between video prediction and pose prediction. Moreover, great success has been made in video prediction \cite{predcnn,fstvp,ld2rvp,sv2p,ptmois,bbsptb} while our pose prediction is rarely researched. Therefore, this motivates us to pursue a dense representation of the 3D skeletal data to smooth this gap.

\begin{figure}[!t]
\centering
\subfloat[]{\includegraphics[width=2.5in]{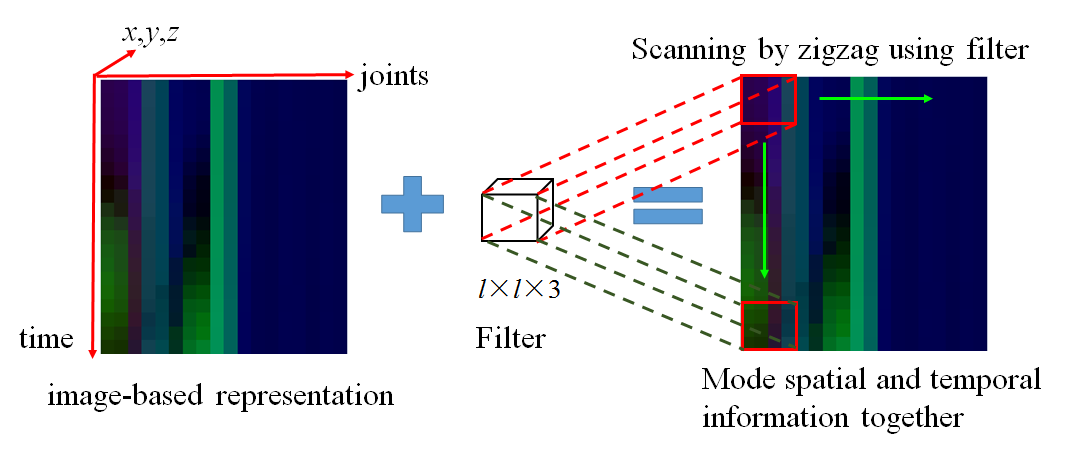}
\label{temporal_evol1}}
\hfil
\subfloat[]{\includegraphics[width=\linewidth]{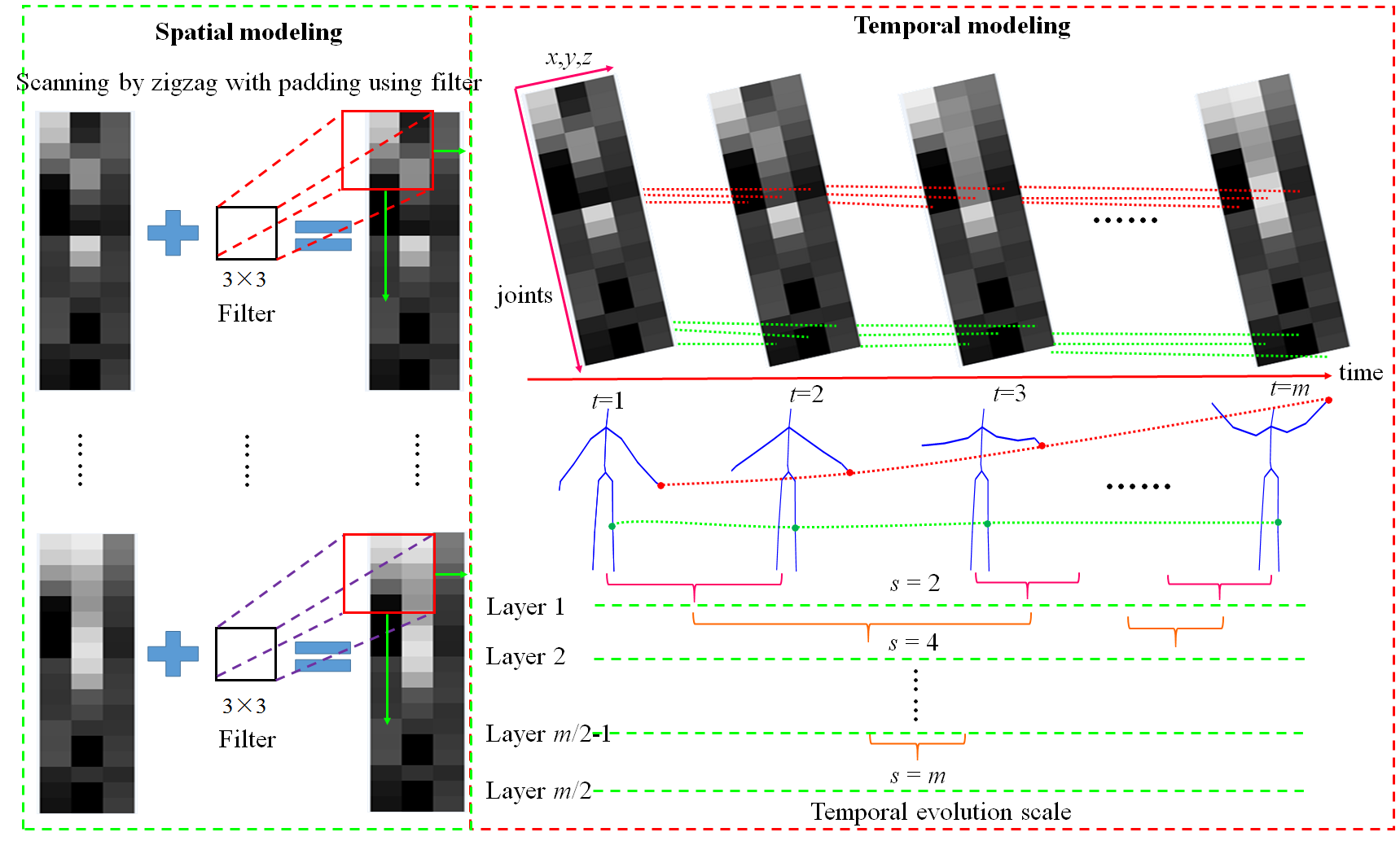}
\label{temporal_evol2}}
\caption{Temporal evolution of previous frames. (a) Commonly spatio-temporal modeling of most of the existing methods. (b) The spatio-temporal modeling of our model.}
\label{temporal_evol}
\end{figure}

{\bf From the perspective of the methodology}, there are two main problems needed to be solved: skeletal representation, sequence to sequence modeling. (${1}$) {\bf Skeletal representation}: As is shown in Figure \ref{temporal_evol1}, most of the image-based skeletal representations are formulated by transforming a joint coordinate sequence into a fixed size image \cite{Kenew,ChaoHCN,DuSAR,skecnn2017,jdmcnn,rltdsbar,lcrsbar,vihar3dbio}. In these representations, the spatial and temporal information are coupled together, while they are different. Recent works are proposed to model the spatial and temporal information separately, and achieve great performance significantly \cite{timeception,tscnn,rstfl,arnvc,s21nar}. Therefore, as is shown in Figure \ref{temporal_evol2}, by decoupling the ``time'' dimension information, we propose a new representation by transforming a joint coordinate sequence into an image sequence, which can conveniently model the spatial and temporal   information differently. Here, ${m}$ is the number of previous frames. At layers ${m}$/2, we can capture the global temporal evolution information. Inspired by \cite{srnnar2015,biodsf2013}, the human body is naturally divided into five parts. Moreover, during the process of human motion, the joints of each part may have strong correlations, and the joints of different parts may have weak correlations. For example, the joints of the left upper limb may have strong correlations, and the joints between upper limbs and lower limbs may have weak correlations. Therefore, to model those different correlations, our new skeletal representation is formulated by ensuring that the joints of each part in the adjacent positions, and the upper limbs and lower limbs in the distant positions separated by the trunk. With this image based skeletal representation, we model the pose prediction problem as the image sequence evolution. (${2}$) {\bf Sequence to sequence modeling}: the state-of-the-art sequence to sequence models predict multiple future frames recursively, which easily suffer from costly computation and error accumulation \cite{predcnn,MartinezOMP,Guifewshot,predrnn}. Moreover, the model complexity increases significantly with the increased length of input or output frames, which is time-consuming. To address those problems, a new framework, PISEP${^2}$, is proposed to predict various future frames in a non-recursive manner by decoupling all the decoders.

In this paper, we present to predict human poses based on the joint coordinate sequence directly. To capture the diverse correlations of different limbs and the local characteristic of the human body, we propose a new skeletal representation by encoding the joint coordinate sequence into the image sequence, which conveniently models the spatial and temporal information of previous frames separately. To efficiently predict future dynamics, we propose a new framework to predict all future frames only in one step. Finally, our method is evaluated comprehensively on G${3}$D \cite{g3d} dataset and FNTU dataset (collected from NTU RGB+D \cite{AmirNTU}), and achieves the state-of-the-art performance.
The main contributions of this paper are summarized as follows:

${1}$. To the best of our knowledge, this is the first time the problem of pose prediction using ${3}$D joint coordinate sequence has been explored, which is intuitive and efficient to evaluate and visualize its performance.

${2}$. A new skeletal representation that preserves the local characteristic of the human body is proposed, which can conveniently model different correlations of different limbs, and also model the spatial and temporal information separately. With this new representation, the pose prediction is modeled as an image sequence evolution problem.

${3}$. A new sequence to sequence model that models the spatial and temporal information differently, PISEP${^2}$, is proposed to predict multiple future frames in one step, which can avoid error accumulation.

${4}$. The proposed method achieves the state-of-the-art performance on two challenging datasets, which shows strong ability of generalization.

\section{Related Work}
Pose prediction receives growing interest recently. In this section, we review the related works from two folds: video prediction and pose prediction.

\subsection{Video prediction}

Video prediction has been extensively studied in recent years \cite{dpcvpul,ostdfvp,fdtvp,pfuvp,dmspms}, but is an unsolved problem. The main issue of video prediction is spatio-temporal modeling. In the following, we review from these perspectives: spatio-temporal modeling using CNN (Convolution Neural Network), spatio-temporal modeling using ConvLSTM (Convolutional Long Short-Term Memory), spatio-temporal modeling using CNN and LSTM.

{\bf Spatio-temporal modeling using CNN}: many approaches were proposed to address the spatio-temporal modeling using CNN in video prediction \cite{acvpag,dmspms,dstc2fp,dpstd,predcnn}. Traditional CNN have shown its power in spatial modeling, and can't efficiently model the temporal information. Therefore, Oh et al.\cite{acvpag} and Zhang et al.\cite{dstc2fp} proposed to concatenate the previous frames along with the axis(i.e. time interval or channels) as one tensor, and then apply a CNN based module to capture the spatio-temporal information. Xu et al. \cite{predcnn} proposed to model the spatial information of each frame by a CNN based block (Residual Multiplicative Block, RMB), and capture the temporal evolution of previous frames hierarchically by a cascade multiplicative unit (CMU) that receives two consecutive frames as input. Zhang et al.\cite{dpstd} proposed to extract spatio-temporal information by convolutions over sequences of tensors.

{\bf Spatio-temporal modeling using ConvLSTM}: ConvLSTM \cite{convlstm} is a special structure of LSTM that combines the effectiveness of CNN and RNN, which can model the spatial and temporal information simultaneously. \cite{ulpivp,predrnn,predrnn2} used ConvLSTM to capture the spatio-temporal information of previous frames for video prediction. For example, Finn et al. \cite{ulpivp} used CDNA (Convolutional Dynamic Neural Advection), a ConvLSTM based framework, to estimate the distribution in the previous frames for each pixel in the new frame, and then used CDNA kernels to predict the motion information. However, conventional ConvLSTM with a layer-independent memory mechanism ignores the memorized information in the previous layers, which is important to predict video sequences. Therefore, Wang et al. \cite{predrnn} proposed a novel framework based on ConvLSTM, ST-LSTM (Spatiotemporal LSTM), to predict multiple future video frames recursively, which can extract and memory the spatial and temporal information simultaneously.

{\bf Spatio-temporal modeling using CNN and LSTM}: Most of the existing works were proposed based on CNN and LSTM to extract spatio-temporal features to predict future video frames recursively \cite{acvpag,dmf2vp,savp,sv2p,fstvp,spvp,dgvp,dpcvpul,vpn,ld2rvp,dmcnvsp}. The commonly modeling of these models are two folds: CNN+ConvLSTM+CNN/${3}$D CNN, CNN+ConvLSTM+Deconvolution. (${1}$) CNN+ConvLSTM+CNN or ${3}$D CNN \cite{savp,sv2p,fstvp,spvp,dgvp,dpcvpul,vpn}: among which, CNNs were commonly used to model the spatial information of previous frames, and ConvLSTMs were used to model the temporal dynamics with local spatial information of the previous frames. Finally, another CNN was applied to predict future video frames. For example, Kalchbrenner et al. \cite{vpn} proposed to model the spatial information with RMBs  of each frame, then used the ConvLSTM to model the dynamic information of previous frames, finally used another RMB to restore the spatial information of future frames. (${2}$) CNN+ConvLSTM+Deconvolution \cite{acvpag,dmf2vp,ld2rvp,dmcnvsp}: in these models, CNN was used to model the spatial information of previous frames and ConvLSTM was used to capture the spatio-temporal of previous frames similarly. Differently, deconvolution was used to predict future frames. For example, Liang et al. \cite{dmf2vp} used VAE (variational autoencoder) to model the distribution of input frames, and then applied ConvLSTM to model the temporal dynamics of previous frames. Then used five deconvolutional layers to predict future frames.

\subsection{Pose prediction}

The related works for pose prediction mainly include: video data based pose prediction, mocap data based pose prediction.

{\bf Video data based pose prediction}: there are lots of works related to the pose prediction based on video data \cite{CaiDVG,pkvf,pfdmm,PFN3D,sp2dnn,mdmfp,rmpof}, which aim to predict human pose sequence with a few image frames. Most of these works used CNN to model the spatial information from the image data, and then used LSTM or DMM to model their temporal dynamics of previous representations or future pose representations \cite{pkvf,pfdmm,PFN3D,rmpof}. For example, \cite{sp2dnn} proposed to predict 3D pose from a static image using CNN based framework to learn the latent pose representation form the static image. \cite{PFN3D} proposed to model the spatial information of the static image, and then used LSTM to model temporal dynamics of the futures. Both \cite{PFN3D} and \cite{sp2dnn} used MSE loss to optimize their models.

{\bf Mocap data based pose prediction}: many works were proposed to predict human poses based on mocap data \cite{ButepageDRL,Guifewshot,MartinezOMP,GuiAdversarial,TangLMP,tprnn,rnmhy,srnnap}. The input and output of these models are mocap vector parameterized by the exponential map \cite{TaylorMM,ppruem}, which is different from a joint coordinate sequence. Most of these works are based on recursive structure, and the model complexity increases with the predictive length of the future pose \cite{GuiAdversarial,TangLMP,rnmhy,srnnap,MartinezOMP,tprnn}. Moreover, most of these models mainly focus on temporal evolution modeling of the previous poses \cite{tprnn,MartinezOMP,rnmhy,srnnap}, which ignores the spatial modeling of the human body. For example, Gui et al.\cite{GuiAdversarial} proposed to predict future human poses using the encoder-decoder network constructed by GRU (Gated Recurrent Unit), then used a residual connection to model the motion velocities of the previous poses. One similar work is based on mocap converted data \cite{ButepageDRL}. The author first covert the mocap frame to the coordinate space in Cartesian coordinates to get a standardized body model. And then designed different encoding-decoding networks based on a fully connected network that model the spatial and temporal information equally, which ignores the differences between the spatial and temporal information.

Different from all previous works discussed above, we formulate a new problem of 3D pose prediction with a 3D joint coordinate sequence instead mocap frame which can be acquired cheaply and efficiently. Moreover, to address this new proposed problem, we propose: (${1}$) a new skeletal representation, which models the problem of pose prediction as the video prediction; (${2}$) a new framework, PISEP${^2}$, to predict all future frames in non-recursive manner, which can avoid error accumulation and improve computational efficiency.

\section{Methodology}
\begin{figure}[!t]
\centering
\includegraphics[width=2.5in]{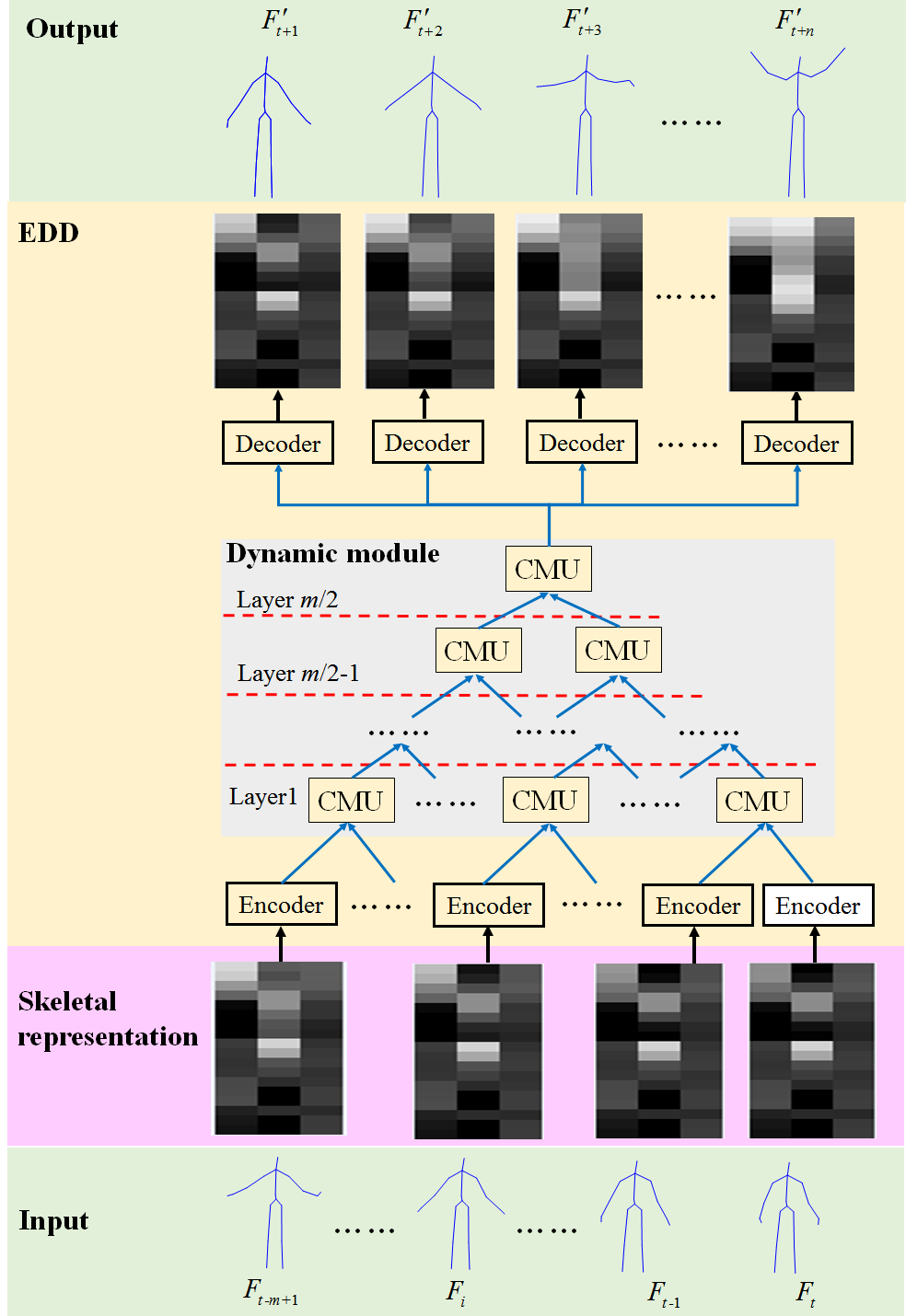}
\caption{Our framework.}
\label{pisepp}
\end{figure}

Our main framework (Image Sequence Evolution based Pose Prediction, PISEP${^2}$) is shown in Figure \ref{pisepp}. It mainly includes four parts: (${1}$) {\bf Input}: our input is the joint coordinate sequence. (${2}$) {\bf Skeletal representation}: the skeletal representation aims to transform the joint coordinate sequence into an image sequence. And the detail of the skeletal representation as described in the following section. (${3}$) {\bf Encoder-Dynamics-Decoder (EDD)}: the goal of this phase is to infer future dynamics through history poses. For this, EDD described in the following is applied. (${4}$) {\bf Output}: finally, the output is the future joint coordinate sequence.

Therefore, in the following sections. We first describe our skeletal representations in detail. Then, we develop a novel framework, EDD, to infer the future dynamics with previous frames. Finally, we briefly introduce the loss function.

\subsection{Skeletal Representation}
\begin{figure}[!t]
\centering
\includegraphics[width=\linewidth]{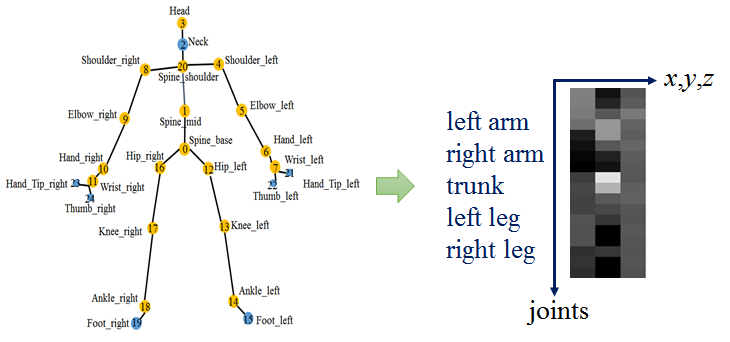}
\caption{The representation of skeletal data. The left part of the figure is the skeleton of the human body and the right part of the figure is the transformational images. }
\label{ske_new}
\end{figure}

Different from image data, the skeleton sequence is a set of joints coordinates. Therefore, in this section, as is shown in Figure \ref{ske_new}, we propose to represent the joint coordinate sequence with the image sequence, which is formulated by preserving the local characteristic of the human body. The left part of Figure \ref{ske_new} is the skeleton of the human body. The numbers in the circle are the index of the skeletal joints. The yellow-colored joints are informative enough for representing human motion \cite{BIPOD}, which is relatively stable. And the blue-colored joints may have a limited effect on the sequence of the poses. Therefore, we will select the yellow-colored joints to formulate the skeletal motion frame.

Given the skeleton of a person in frame ${i}$, this new representation is represented by transforming the joints coordinates ${S_i}$ into a one-channel image ${F_i}$ denoted as equation \ref{eqn1}. Where ${S_i} = \{{J_1},{J_2}, \cdots ,{J_N}\}$, ${J}$ = (${x} $,${y}$,${z}$)and ${N}$ is the number of joint. With this representation, the joint coordinate sequence is represented by an image sequence: $\{ {S_1},{S_2}, \cdots ,{S_m}\}  \Rightarrow \{ {F_1},{F_2}, \cdots ,{F_m}\}$,where ${m}$ is the length of a sequence. Besides, motivated by \cite{srnnar2015,biodsf2013}, the human body can be divided into five parts: left arm, right arm, trunk, left leg, and right leg. To model the different relationships of different limbs, as shown in the right part of Figure \ref{ske_new}, we propose to: (${1}$) place the two limbs or two legs in the adjacent areas, which can conveniently model the correlation of two limbs or two legs; (${2}$) place the two limbs and the two legs in the distance areas separated by trunk, which can efficiently capture their weak correlations. Moreover, to maintain the local characteristic, we place the joints of each limb in the continuous adjacent areas, which keep the physical connection relationship of the local joints. Then, the joints of the five parts are concatenated in the order: left arm, right arm, trunk, left leg, right leg. Finally, our joints orders are: ${11,10,9,8,4,5,6,7,3,20, 1,0,16,17,18,12,13,14}$. Different from \cite{srnnar2015}, our pose is represented by a two-dimension matrix that preserves the local structure of the human body, while \cite{srnnar2015} represents their pose with a one-dimension vector by orderly concatenating the ${x}$,${y}$,${z}$ coordinates of each joint that loses the local structure of the human body to some extent.

\begin{equation}
\label{eqn1}
{F_i} = \left[ {\begin{array}{*{20}{c}}
{{x_1}}&{{y_1}}&{{z_1}}\\
{{x_2}}&{{y_2}}&{{z_2}}\\
 \vdots & \vdots & \vdots \\
{{x_N}}&{{y_N}}&{{z_N}}
\end{array}} \right]
\end{equation}

The right image in Figure \ref{ske_new} is the transformational pseudo image. The size of the transformed images is ${18 \times 3}$. Here, in our dataset, the value of the ${x}$ or ${y}$ coordinate is very small, and the value of the ${z}$ coordinate is relatively large. Therefore, we translate the ${x}$, ${y}$, and ${z}$ coordinate values appropriately to obtain a better display effect. The same processing is applied in Figure \ref{temporal_evol} and Figure \ref{pisepp} for a better visual presentation. In the skeletal representation, the pixel value represents the ${x}$/${y}$/${z}$ coordinates of the corresponding joint, and the change of the pixel value represents the movement of the joint in the ${x}$/${y}$/${z}$-axis direction. Notably, because the coordinate value may be affected by the distance of the position of the camera, the coordinate value may exceed the effective representation of the image. Therefore, we can force the coordinate value of joints in the effective representation range of the image without changing the pose shape and the temporal evolution process of pose sequence through uniform translation and scale scaling operations. In this paper, since our joints coordinate values are far from this range, we have not made any processing in all experiments. Notably, the pixel values are real numbers, including positive real numbers, negative real numbers and ${0}$, which is consistent with the joint coordinate values, and different from the general image.

The advantages of our skeletal representation are summarized as follows:

${1}$.	{\bf Modeling the different correlations of joints of the human body}: On one hand, by ensuring the joints of each limb in the adjacent areas, we can conveniently model the strong correlations of joints of each limb. On the other hand, by placing the upper limbs and the lower limbs in the distant areas, we can conveniently model their weak correlations.

${2}$.	{\bf Modeling the spatial and temporal information of previous frames separately}: as shown in Figure \ref{temporal_evol2}, by decoupling the ``time'' dimension, we can conveniently model the spatial and temporal information separately.

\subsection{EDD}
\begin{figure*}[!t]
\centering
\includegraphics[width=6in]{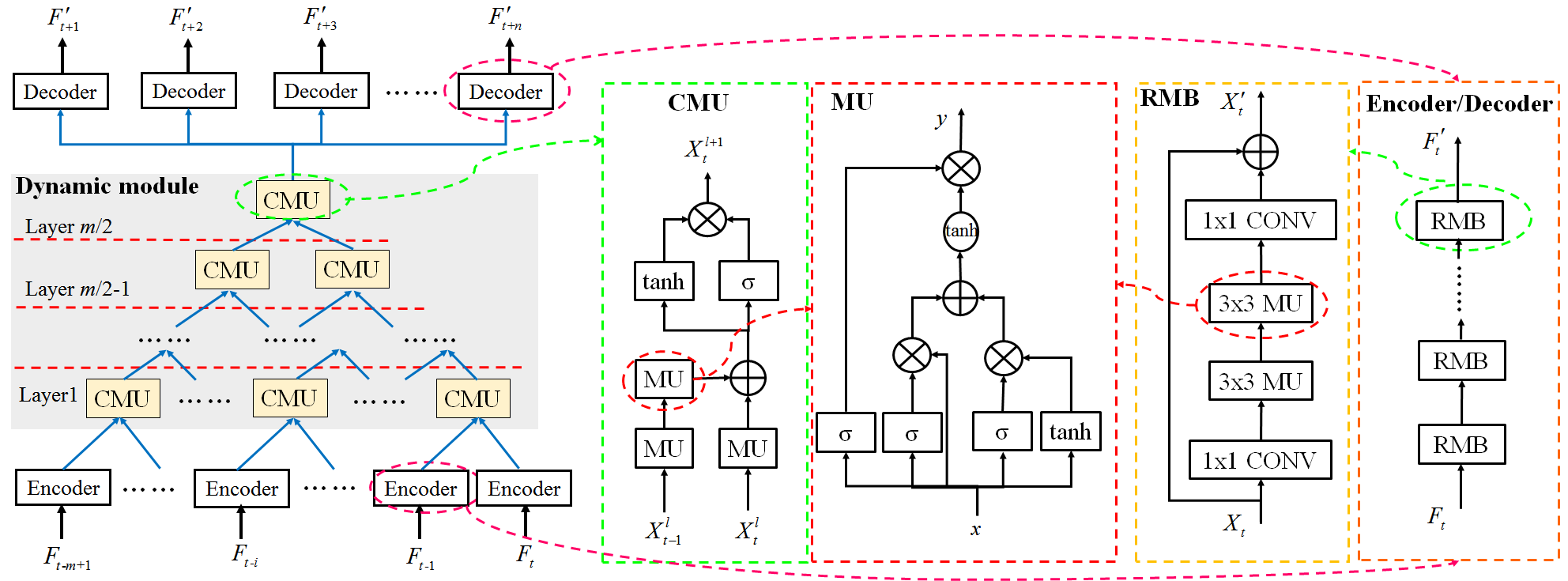}
\caption{The framework of EDD.}
\label{edd}
\end{figure*}

Most existing sequence-to-sequence models likely suffer from costly computation and error accumulation with its recurrent structure \cite{predcnn,MartinezOMP,Guifewshot,fstvp}. Inspired by the architecture of \cite{predcnn} for spatio-temporal modeling, we propose a new structure, EDD (Encoder-Dynamics-Decoder), to predict multiple future frames in one step, which can significantly improve the computational efficiency and avoid error accumulation. Figure \ref{edd} is the framework of EDD. The framework mainly includes three parts: (${1}$) {\bf Encoder}: this module aims to model the spatial structure information of the pose. Residual multiplicative block (RMB) \cite{vpn} has power for spatial modeling with its LSTM-like structure. Therefore, we introduce RMB as the basic unit. The encoder is stacked by ${l_e}$ residual multiplicative blocks (RMBs) to enlarge its receptive field for better spatial modeling. To reduce the model complexity, the encoders share weights. (${2}$) {\bf Dynamics module}: the dynamic module aims to capture the general temporal evolution of previous poses. For this, we introduce the cascade multiplicative unit (CMU) proposed in \cite{predcnn} as the building block which models the dynamics of the adjacent frames. To capture the global temporal evolution information of previous frames, we model the different scales of temporal information in a hierarchical way. More specially, ${2^l}$ continuous frames temporal dynamics will be captured at layer ${l}$ in the dynamics module. At layer ${m}/2$, we can model the global temporal information of previous ${m}$ frames. Therefore, we can model the different scales of temporal evolution information at different layers. To reduce the computational cost, CMUs at each layer share weights; to capture different scales temporal information, CMUs at different layers are decoupled with each other. Different from \cite{predcnn}, we remove the chain structure of this model module to learn the general temporal evolution information only once at all time steps, while \cite{predcnn} need to learn the temporal evolution of previous frames recursively at each step. (${3}$) {\bf Decoder}: this module proposes to predict all future frames in one step. The decoder is stacked by ${l_d}$ RMBs, which reconstruct the spatial structure of predictive pose. Different future poses are generated by decoupling all the decoders with each other based on the general temporal evolution information produced by dynamics module. Differently, all decoders in \cite{predcnn} share weights.

\subsection{Loss function}
Our goal is to predict future poses as close as possible to groundtruth. ${L2}$ norm loss is the commonly used loss function for the similar problem. Obviously, when the number is smaller than one, its square value smaller than absolute. In this case, the L1 norm loss can better reflect the difference between these two similar poses. Therefore, we propose to achieve a more accurate pose prediction using ${L1}$ norm loss that may better guide training well. Our loss function can be formulated as equation \ref{eqn2}:

\begin{equation}
\label{eqn2}
l = \left\| {y - \widehat y} \right\|\
\end{equation}
Where ${y}$ is the groundtruth pose, ${\widehat y}$ is the predictive pose. During training, we aim to minimize the loss function to obtain the optimization results.

\section{Experiments}

In this section, we evaluate our model on two challenging datasets: G3D \cite{g3d} and FNTU. Where FNTU dataset is collected from NTU RGB+D \cite{AmirNTU} dataset to ensure the quality of the learning data. We first introduce the datasets and implementation details. Then, we briefly explain our baselines. Next, we compare our method with the state-of-the-art methods to verify the performance of PISEP${^2}$. And extensive experiments are conducted to evaluate the effectiveness of our proposed method. Moreover, we further evaluate the performance of our model for unseen data. Finally, we show some qualitative results of pose prediction.

\subsection{Dataset and Implementation Details}

{\bf G${3}$D}: G${3}$D \cite{g3d} dataset contains ${10}$ subjects performing ${20}$ gaming actions captured by the Microsoft Kinect sensor. G${3}$D is an unsegmented dataset, and each video may contain multiple actions. It consists of ${210}$ samples in total. We randomly select ${70}$ samples as a test set and the rest as a training set.\footnote {\label{dataset} Our datasets: https://drive.google.com/drive/folders/1bqNyIk2O0NIf5Hv
2sMfwsuPjwbpZK-n5}

{\bf Filtered NTU RGB+D (FNTU)}\footref{dataset}: NTU RGB+D \cite{AmirNTU} dataset is collected by the Microsoft Kinect v${2}$ sensor. The dataset includes 60 action classes performed by ${40}$ subjects and consists of ${56,880}$ video clips in total. NTU RGB+D dataset is a well-segmented dataset, and each video contains one action. However, the skeletal data captured from the side or back of the human body is very noisy. And those noisy skeletal sequences are not suitable for pose prediction. Therefore, we form our dataset based on NTU RGB+D dataset, named filtered NTU RGB+D (FNTU), by: (${1}$) filtering the mutual actions since our method focus on the analysis of single person movement of the human body; (${2}$) selecting the relative forward skeleton of the human body to ensure the quality of learning data. The FNTU dataset consists of ${18102}$ samples. We randomly select one of ${12001}$ samples for training and the rest for testing.\footref{dataset}

{\bf Implementation Details}: in experiments, we use an overlap sliding window to clip the skeleton sequence on the training set and test set respectively. We aim to use the previous ${10}$ frames to predict future ${10}$ frames. Therefore, the window size is set to ${20}$. To ensuring the continuity of the segmented sequence, the overlap size is set to ${5}$. Our final training set contains ${3543}$ sequence clips, and the test set contains ${1637}$ sequence clips on the G${3}$D dataset. And the training set consists of ${53843}$ samples, and the test set consists of ${26819}$ samples on FNTU. To avoid over-fitting, we stack ${2}$ RMBs as the encoder and stack ${3}$ RMBs as the decoder on G${3}$D dataset. And we stack ${4}$ RMBs for the encoder and stack ${6}$ RMBs for the decoder on FNTU dataset. We train all models using Adam optimizer, and our learning rate is initial with ${0.0001}$. All the experiments are implemented with TensorFlow.

{\bf Metrics}: We choose the mean squared error (MSE) per frame and the mean absolute error (MAE) per frame as our evaluation metrics. Specially, as shown in equations \ref{eqn3} and \ref{eqn4}, we calculate the MSE or MAE between the joint coordinate of the groundtruth pose and the joint coordinate of the predictive pose and normalize with the length of the predictive sequence.

\begin{equation}
\label{eqn3}
{e_{mse}} = \sum\limits_i^N {\sum\limits_j^3 {{{({p_{i,j}} - {{\widehat p}_{i,j}})}^2}} }
\end{equation}

\begin{equation}
\label{eqn4}
{e_{mae}} = \sum\limits_i^N {\sum\limits_j^3 {|{p_{i,j}} - {{\widehat p}_{i,j}}|} }
\end{equation}
Where ${p}$ and ${\widehat p}$  represented by equations (${1}$) are the groundtruth pose and predictive pose respectively, ${N}$ is the number of joint, ${p_{i,j}}$ is the value of the ${i}$th joint, ${j}$th dimension of the groundtruth pose, and ${\widehat p_{i,j}}$ is the value of the ${i}$th joint, ${j}$th dimension of the predictive pose.

\subsection{Baselines}
Since pose prediction with a joint coordinate sequence is the newly proposed problem, there exists no baseline for comparison. Our pose prediction is modeled as the problem of image evolution, we consider the PredCNN proposed in \cite{predcnn} as our baseline. Moreover, the most similar work in \cite{ButepageDRL} converts the mocap frame into the joint coordinate frame in Cartesian coordinates. Therefore, for comparison, we reproduce the framework (Symmetric Temporal Encoder, S-TE) as our baseline.

{\bf PredCNN}: the non-overlapping PredCNN is the most similar to ours. For a fair comparison, the non-overlapping PredCNN with our new skeletal representation is introduced to address the problem of our proposed pose prediction.

{\bf S-TE}: as shown in \cite{ButepageDRL}, the framework of S-TE has five fully connected layers, and all layers are of dimensions ${(3 \times {N_{{\rm{joints}}}} \times {N_i}}$ , ${300}$, ${100}$, ${300}$, ${(3 \times {N_{{\rm{joints}}}} \times {N_o})}$ respectively. Where the ${{N_{{\rm{joints}}}}}$, ${N_i}$ and ${N_o}$ are the joints number of a pose, the input length of the previous frames and output length of the future frames respectively. In this paper, the joints number of a pose is ${18}$ (i.e. ${{N_{{\rm{joints}}}} = 18}$). Since the joints order does not affect the network performance with fully connected structure, given a window frame of size ${N}$, we process the data by flattening our skeletal representation described above into a vector with dimensions ${(3 \times {N_{{\rm{joints}}}} \times N)}$.

\subsection{Comparison with baselines}
To evaluate the performance of our framework, we compare our method with the above baselines. The experimental results are shown in Table \ref{table1}. And our model achieves the state-of-the-art performance, which demonstrates the effectiveness of our proposed method. Compared with PredCNN \cite{predcnn}, our method significantly gains in accuracy on both datasets. For example, the MSE decreases from ${0.1882}$ to ${0.1199}$, and the MAE decreases from ${1.5713}$ to ${1.1101}$ on G${3}$D. The experimental results show that our framework can avoid error accumulations. Besides, our model captures the spatial information by an LSTM-like block, and the temporal information with a hierarchical structure, which can treat the spatial and temporal information unequally, while the S-TE \cite{ButepageDRL} treat the spatial and the temporal information equally. Therefore, compared with the S-TE \cite{ButepageDRL}, our framework can handle the spatial-temporal information well. This may be the possible reason that our model leads to slightly better performance. For example, compared with S-TE \cite{ButepageDRL}, the MSE increases by ${0.0208}$ and ${0.0215}$, and the MAE increases by ${0.0994}$ and ${0.1443}$ on G${3}$D and FNTU, respectively.

\begin{table}[!t]
\renewcommand{\arraystretch}{1.3}
\caption{Comparison with the state-of-the-art methods}
\label{table1}
\centering
\begin{tabular}{ccccc}
\hline
\multirow{2}{*}{}& \multicolumn{2}{c}{MSE} & \multicolumn{2}{c}{MAE} \\
%\hline
 \cline{2-5}& G3D &FNTU & G3D &FNTU \\
\hline
PredCNN\cite{predcnn}&0.1882&0.1665&1.5713&1.6394 \\
%\hline
S-TE\cite{ButepageDRL}&0.1407&0.1425&1.2095&1.3094 \\
%\hline
PISEP${^2}$&{\bf 0.1199}&{\bf 0.1210}&{\bf 1.1101}&{\bf 1.1651} \\
\hline
\end{tabular}
\end{table}

\begin{figure}[!t]
\centering
\includegraphics[width=2.5in]{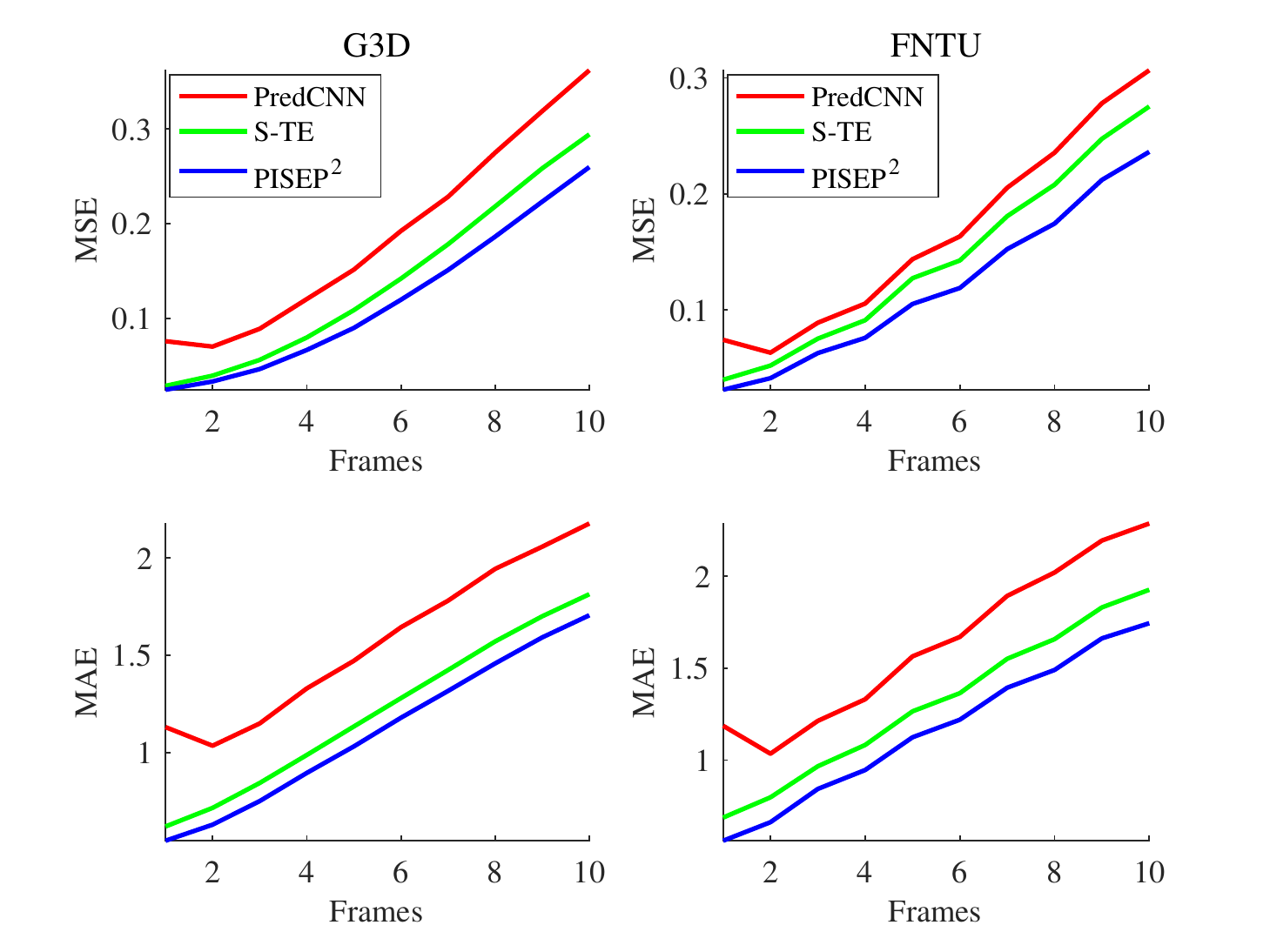}
\caption{Frame-wise performance of different methods.}
\label{fig6}
\end{figure}

{\bf Quantitative analysis of frame-wise results}: to analyze the performance of each time-step, the frame-wise performance of different methods are as shown in Figure \ref{fig6}. Where the horizontal axis represents by frames and the vertical axis represents by MSE or MAE of each frame. The mean MSE of predictive poses for each frame is ${0.1199}$ and ${0.1210}$, and the mean MAE of predictive poses for each pose is ${1.1101}$ and ${1.1651}$ on G${3}$D and FNTU, respectively. Compared with PredCNN \cite{predcnn}, our method significantly decreases error at all time-steps, especially for the long-term prediction. Moreover, PredCNN easily suffers from error accumulation that may lead to its poor performance, and it may more obvious in the later time-step. The experimental results show that our method can significantly enhance the predictive performance on both short-term and long-term predictions. And our framework achieves the best performance, which further evidence the effectiveness of our proposed method.

\begin{figure}[!t]
\centering
\subfloat[]{\includegraphics[width=3.85in]{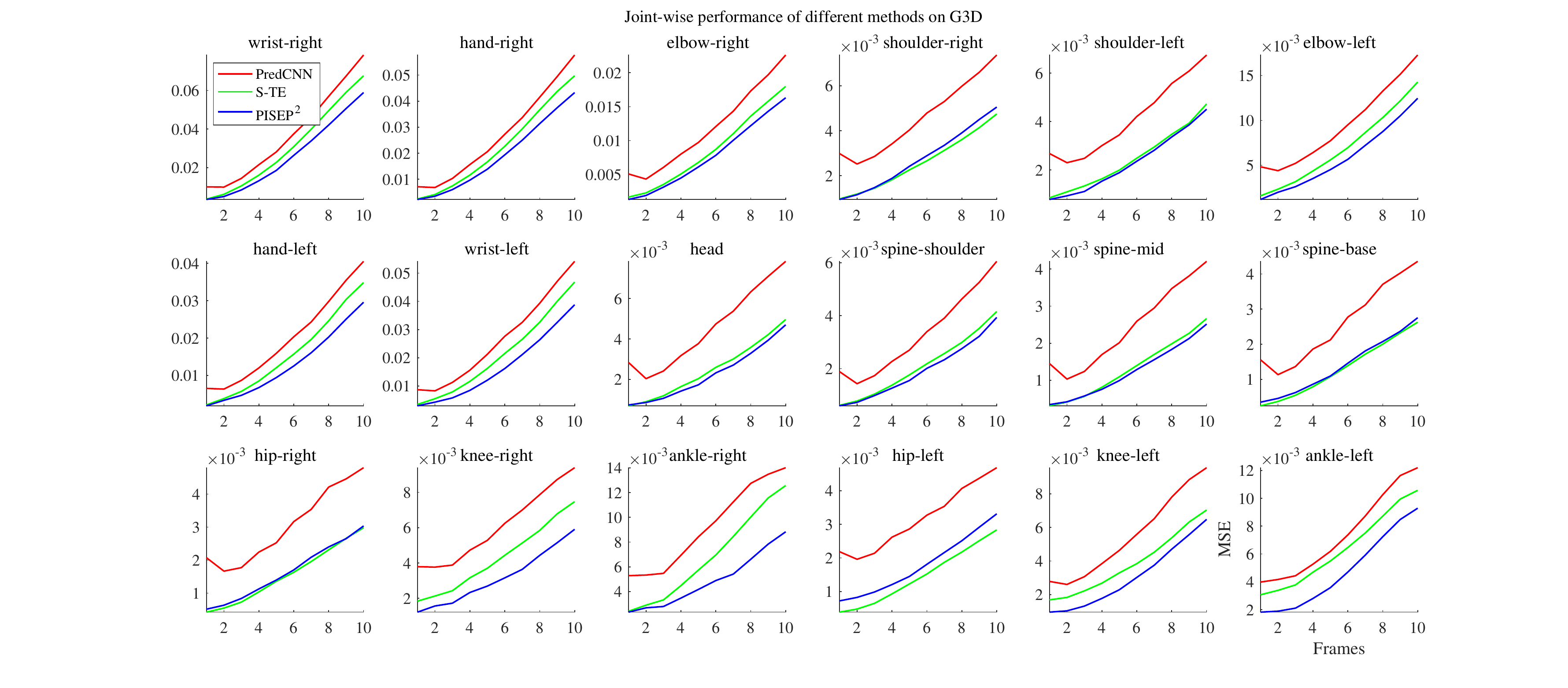}
\label{fig7_1}}
\hfil
\subfloat[]{\includegraphics[width=3.85in]{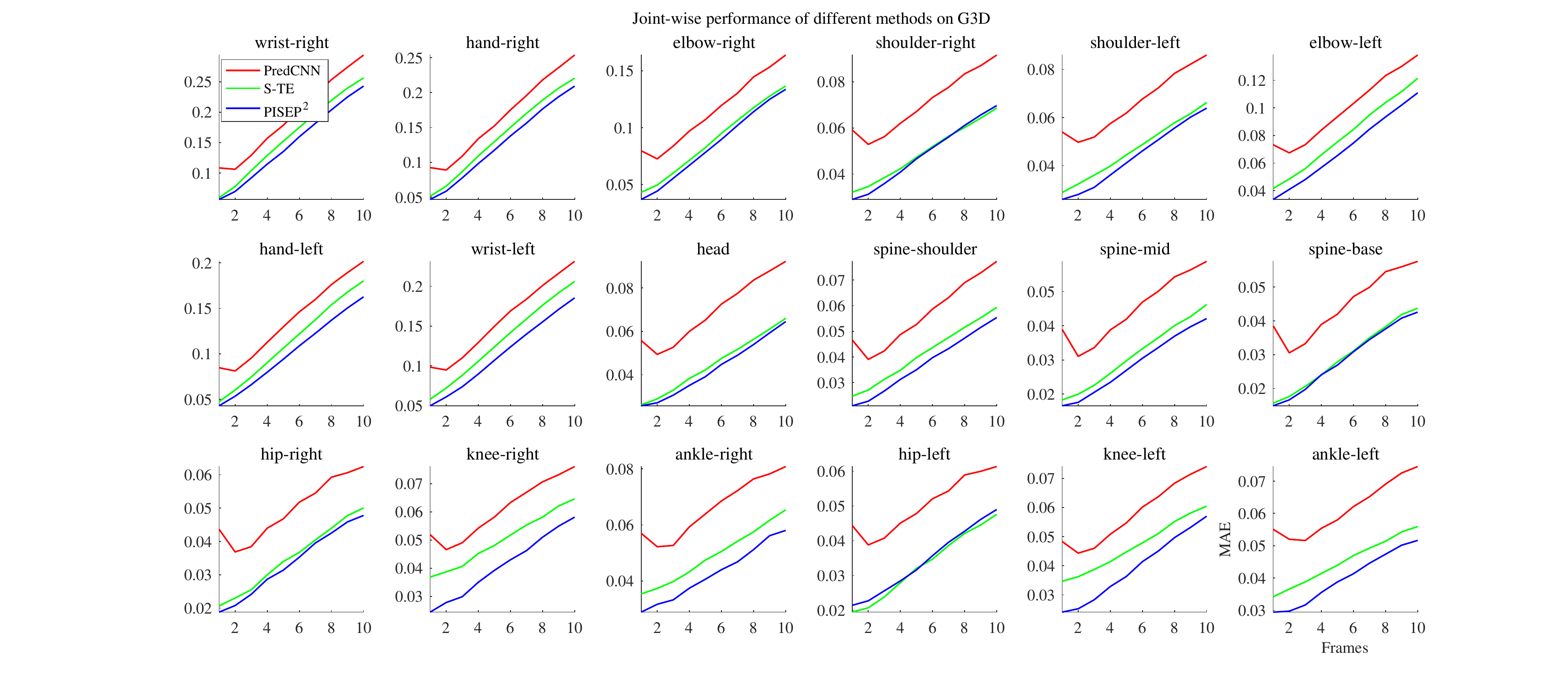}
\label{fig7_2}}
\hfil
\subfloat[]{\includegraphics[width=3.85in]{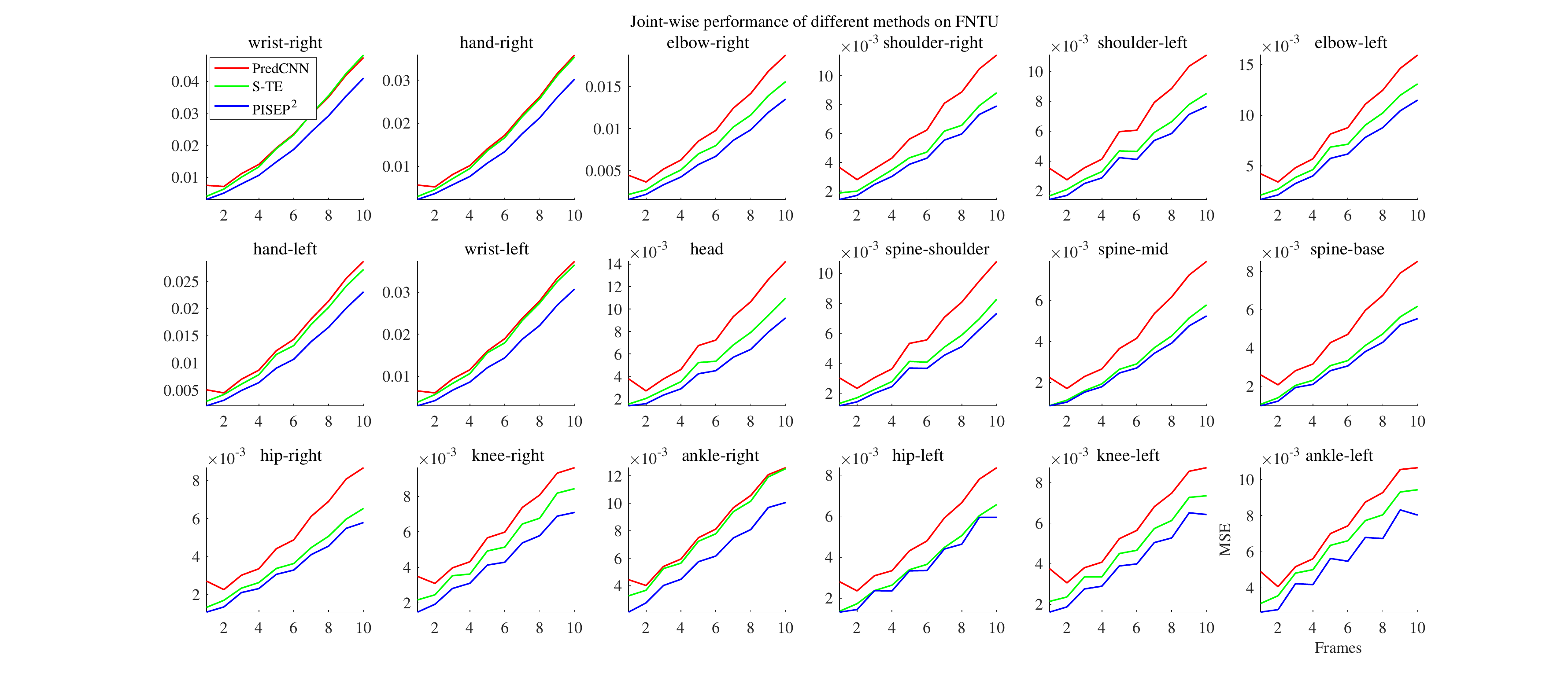}
\label{fig7_3}}
\hfil
\subfloat[]{\includegraphics[width=3.85in]{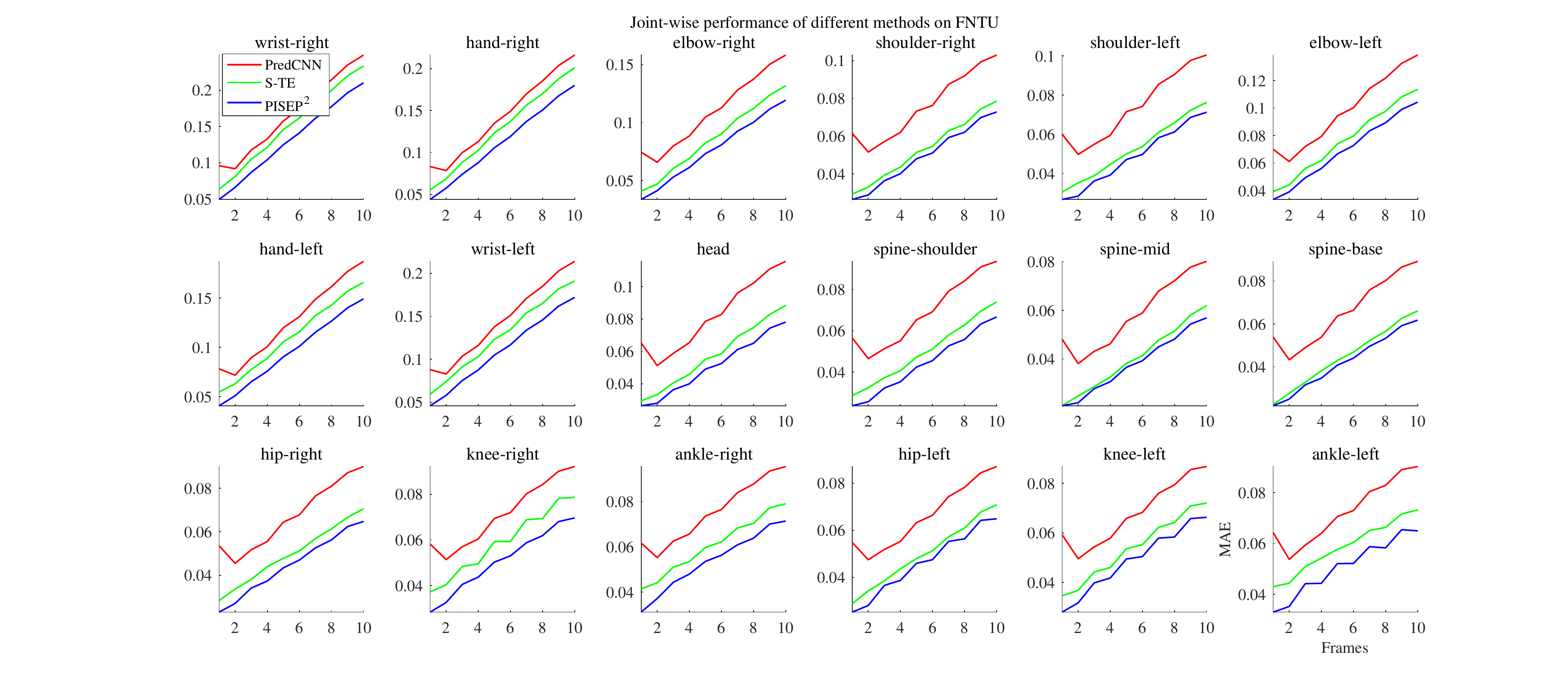}
\label{fig7_4}}
\caption{Joint-wise performance of different methods. (a) Joint-wise MSE of different methods on G${3}$D. (b) Joint-wise MAE of different methods on G${3}$D. (c) Joint-wise MSE of different methods on FNTU. (d) Joint-wise MAE of different methods on FNTU. }
\label{fig7}
\end{figure}

{\bf Quantitative analysis of joint-wise results}: to further analyze the performance of our method, we measure the error of each joint. Figure \ref{fig7} is the joint-wise performance of different methods. Where the horizontal axis represents by frames and the vertical axis represents by MSE or MAE of each of joint. (${1}$) {\bf On G${3}$D}, the errors of the joints of the upper limbs are relatively large, while the errors of the joints of lower limbs or trunk are relatively small. The possible reason for this phenomenon is that most of the movement occurs mainly in the joints of the upper limbs. Compared with the upper limbs joints, the joints of lower limbs or trunk are relatively stable. Therefore, the errors of the joints of the upper limbs are larger than the errors of the joints of the lower limbs or trunk. Compared with PredCNN, the performance of our method is significantly extends their performance at all joints, which demonstrates that our method can avoid error accumulation well. Compared with S-TE, our method outperforms S-TE overall for both MSE and MAE at all joints. Since our model treats the spatial and temporal information differently, while S-TE model the spatial and temporal information equally, our method can better capture the temporal evolution information. And the experimental results have demonstrated this to a great extent. (${2}$) {\bf On FNTU}, compared with PredCNN and S-TE, our method achieves the best results for both MSE and MAE. More specially, the errors of the upper limbs joints are relatively large, and the errors of the joints of the lower limbs or trunk are relatively small, which demonstrates similar results on G${3}$D. Similarly, most of the actions on FNTU are the upper limbs related action. The movement of the upper limbs joints is relatively violent. This may be the main reason for the above phenomenon. The experimental results on FNTU are consistent with the results on G${3}$D, which intensively verify the effectiveness of our proposed method.

\begin{figure}[!t]
\centering
\subfloat[]{\includegraphics[width=3.85in]{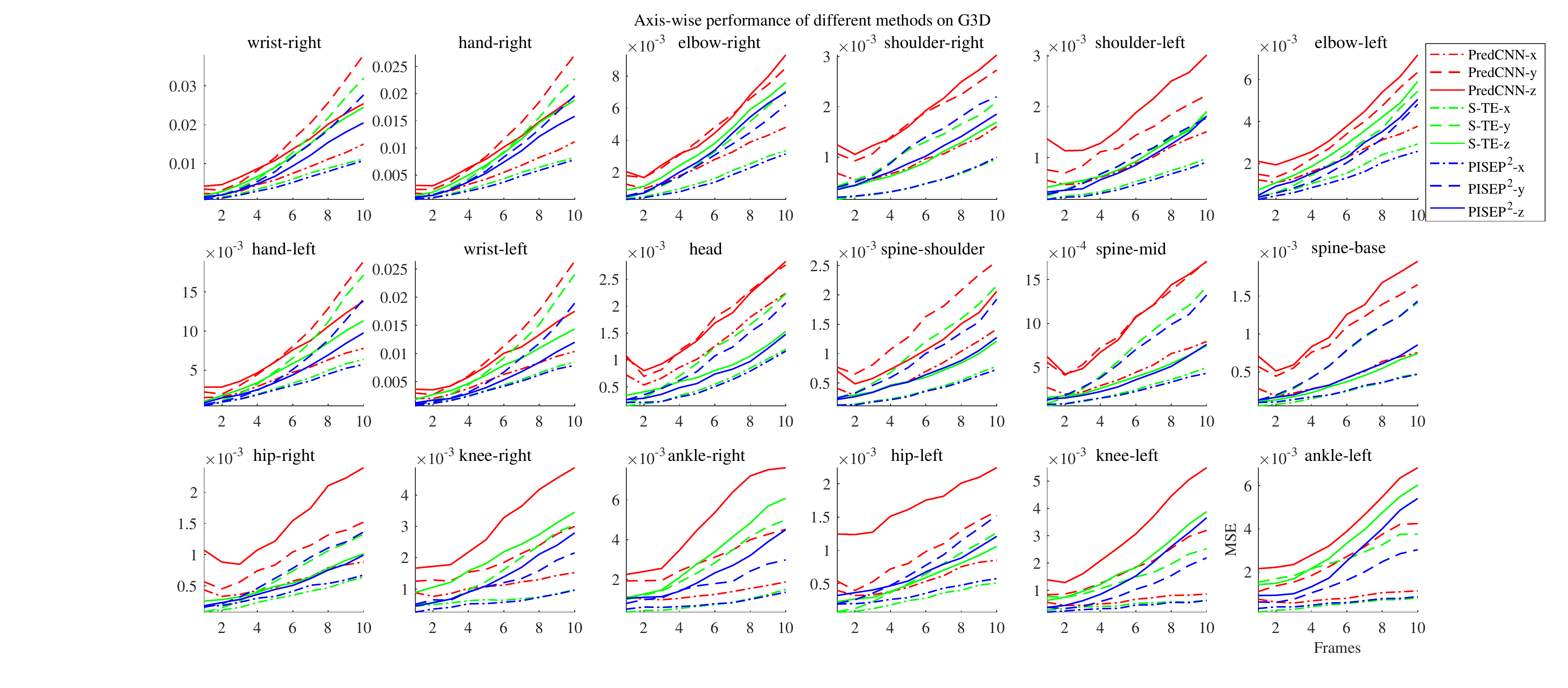}
\label{fig8_1}}
\hfil
\subfloat[]{\includegraphics[width=3.85in]{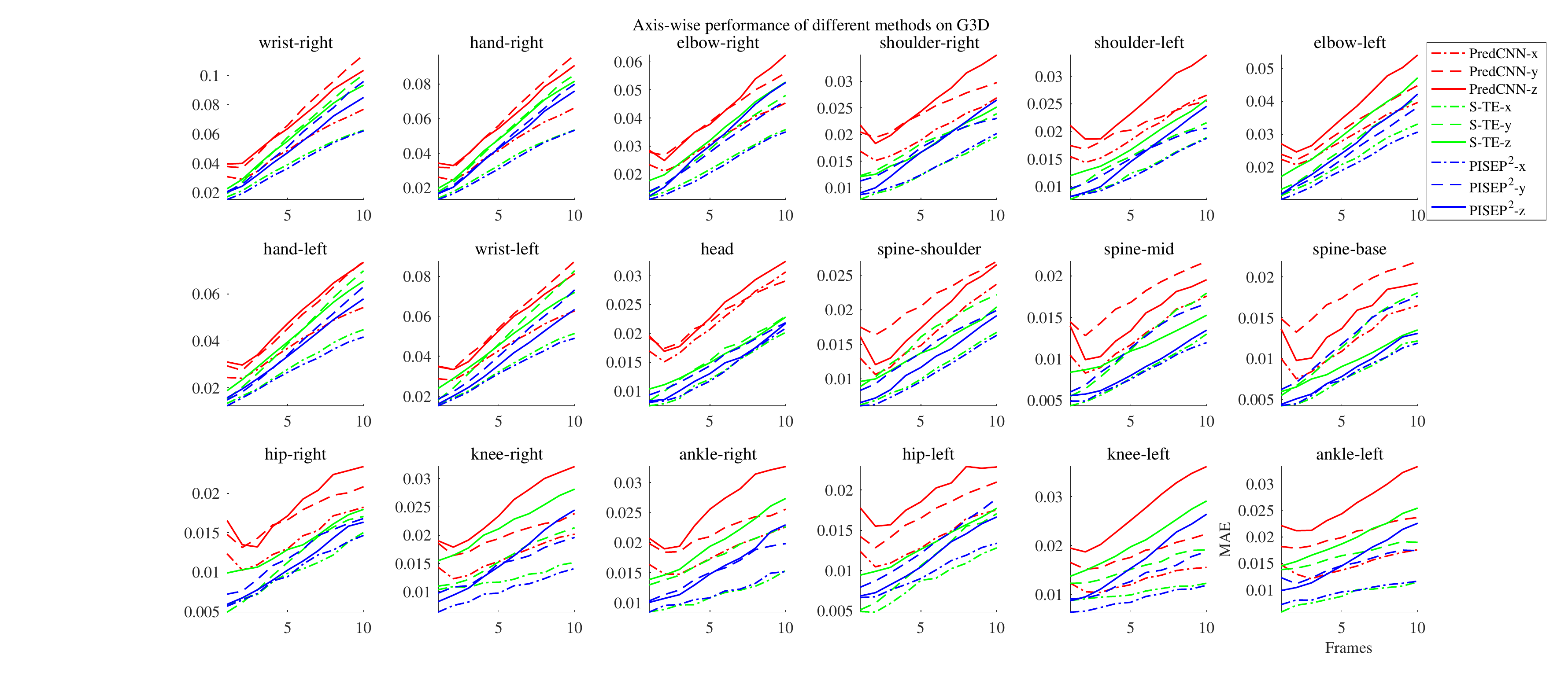}
\label{fig8_2}}
\hfil
\subfloat[]{\includegraphics[width=3.85in]{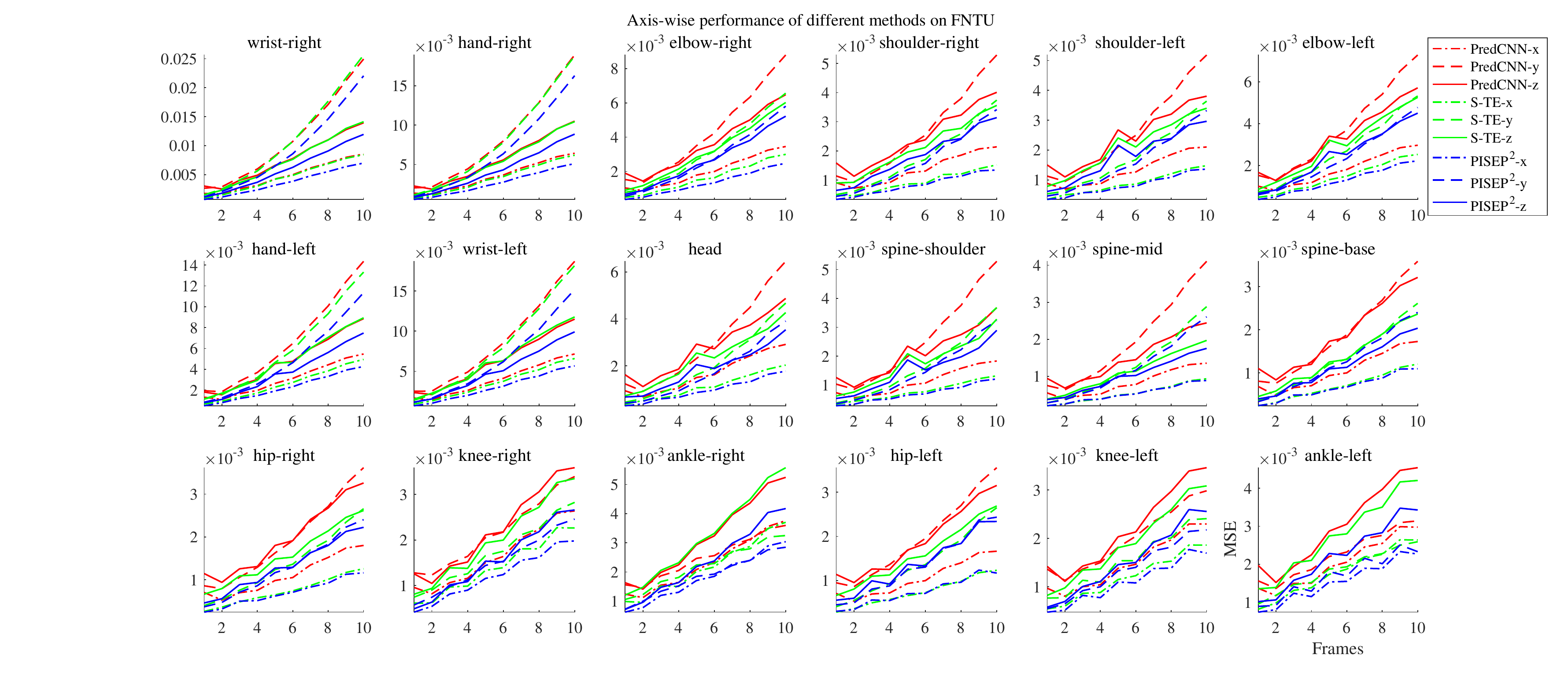}
\label{fig8_3}}
\hfil
\subfloat[]{\includegraphics[width=3.85in]{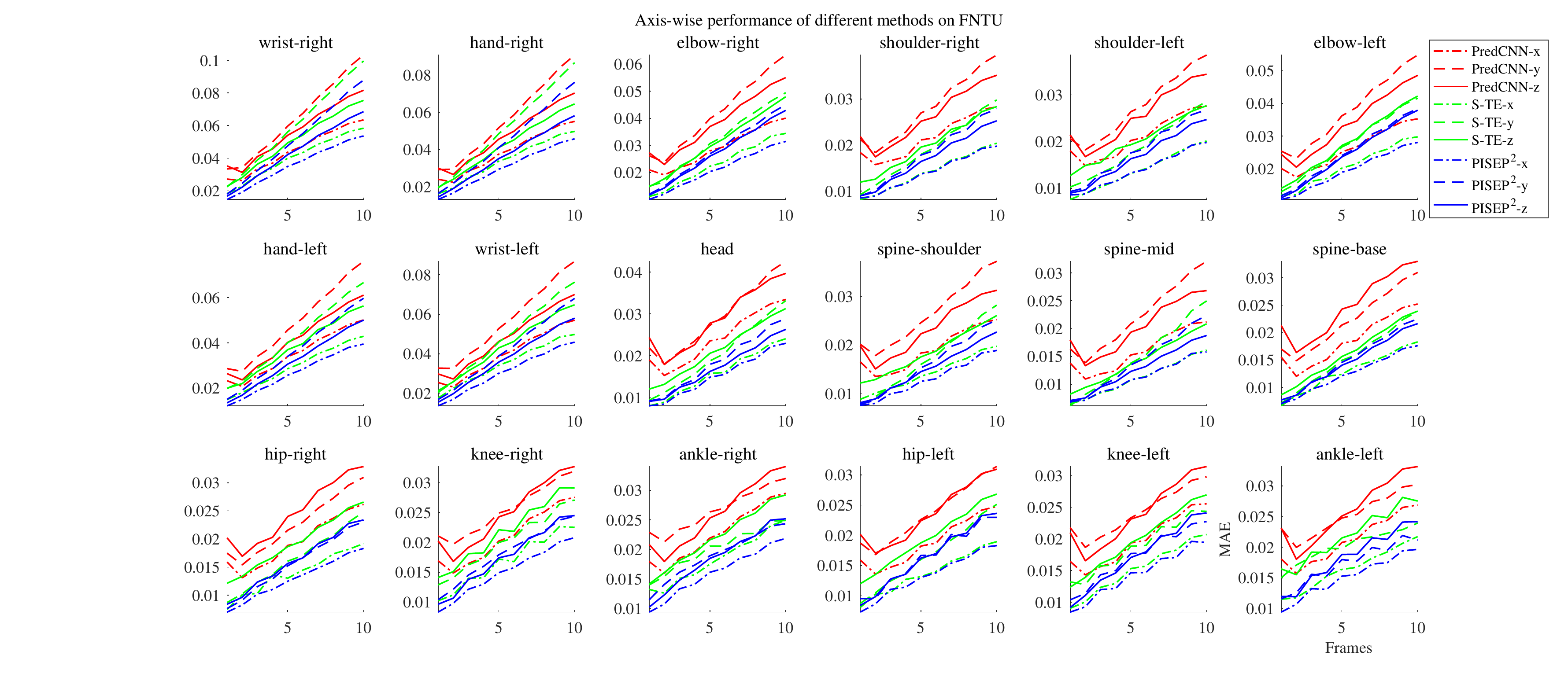}
\label{fig8_4}}
\caption{Axis-wise performance of different methods. (a) Axis-wise MSE of different methods on G${3}$D. (b) Axis-wise MAE of different methods on G${3}$D. (c) Axis-wise MSE of different methods on FNTU. (d) Axis-wise MAE of different methods on FNTU. }
\label{fig8}
\end{figure}

{\bf Quantitative analysis of axis-wise results}: because the ${z}$ coordinate value of each joint is significantly larger than the ${x}$ or ${y}$ coordinate value of each joint, to further analyze the error of each predictive pose, the errors are calculated along the axis ${x}$, axis ${y}$, and axis ${z}$ respectively. Figure \ref{fig8} is the axis-wise performance of different methods. Where ``*-${x}$'' denotes the error of method ``*'' on the axis ${x}$, ``*-${y}$'' denotes the error of method ``*'' on the axis ${y}$, ``*-${z}$'' denotes the error of method ``*'' on the axis ${z}$. (${1}$) {\bf On G${3}$D}: in general, the errors of ${x}$, ${y}$, and ${z}$ coordinate of joints of upper limbs are relatively large, especially for the ``wrist righ'',``hand right'', ``hand left'', and ``wrist lef'' joint, on both MSE and MAE. The possible reason for this is that: the actions on G${3}$D are the upper limbs related actions, and these joints are the most active. Therefore, this may lead to a large error of these joints. For all the methods, generally, the errors of the ${x}$ coordinate of each joint are the smallest. The errors of ${z}$ coordinate of the joints of the upper limbs are relatively large, while the errors of the ${y}$ coordinate of the joints of the lower limbs or trunk are relatively large. The possible reason is: for all actions, in general, (${a}$) the movement of the ${x}$-axis is the smallest for all joints; (${b}$) the movement of the ${y}$-axis of upper limb joints is more violent; (${c}$) the movement of the ${z}$-axis of the trunk or lower limbs joints is more violent. This may cause the smallest errors of the ${x}$ coordinate of each joint, the errors of the ${y}$ coordinate of the joints of the upper limbs may larger than the trunk or lower limbs joints, and the errors of ${z}$ coordinate of the trunk or lower limbs joints may larger than the joints of the upper limbs. Compared with \cite{predcnn}, the errors of our method are decreased significantly for both ${x}$, ${y}$, and ${z}$ coordinate of all joints, which demonstrate that our model can efficiently avoid error accumulation. Compared with \cite{ButepageDRL}, the performance of our method is slightly better, which further demonstrates that our method can handle the temporal evolution of previous frames. (${2}$) {\bf On FNTU}, this shows similar results for both MSE and MAE, which demonstrate the effectiveness of our method again. Specially, the ${x}$ coordinate of all joints achieves the best performance. For the joints of the upper limbs, the errors of the ${y}$ coordinate of each joint are relatively large; for the joints of the trunk or lower limbs, the errors of the ${z}$ coordinate of each joint are relatively large. The possible reason is similar to the results on G${3}$D as discussed above. Compared with \cite{predcnn}, the performance of our method significantly outperforms \cite{predcnn} for all joints on both ${x}$, ${y}$, and ${z}$-axis, especially for the joints of the upper limbs, which demonstrates that our method captures the temporal evolution well and can efficiently avoid error accumulation. Compared with \cite{ButepageDRL}, our method outperforms \cite{ButepageDRL} at all joints overall, especially for the joints of the upper limbs that movement violently, which further shows the effectiveness of our proposed method to capture temporal evolution information of previous poses.

\subsection{Evaluation of PISEP${^2}$}
To further evaluate the effectiveness of our proposed method, in this section, we will comprehensively verify from there three aspects: skeletal representation, network architecture, and loss function.

{\bf Evaluation of skeletal representation}: to evaluate the performance of our skeletal representation, we randomly disrupt the order of the joints to evaluate our network. Table \ref{table2} is the results of different representations. Where ``disorder${1}$'' and ``disorder${2}$'' are two group experiments of random joints order representation. Compared with the disorder joints representation, our method outperforms the performances of ``disorder${1}$'' and ``disorder${2}$'' representation on both G${3}$D and FNTU, which demonstrates our skeletal representation can preserve the local characteristic of the human body. Besides, with our skeletal representation, we can efficiently model the different correlations of different limbs. However, the improvement of our skeletal representation is limit. For example, the MSE only decreases by ${0.0033}$ or ${0.0064}$, and the MAE decreases by ${0.0180}$ or ${0.0437}$ on G${3}$D. The possible reason is that our new representation is too small, and, to some extent, different joints may be affected by each other under the operation of convolution.

\begin{table}[!t]
\renewcommand{\arraystretch}{1.3}
\caption{Results of different representations}
\label{table2}
\centering
\begin{tabular}{ccccc}
\hline
\multirow{2}{*}{Representation}& \multicolumn{2}{c}{MSE} & \multicolumn{2}{c}{MAE} \\
 \cline{2-5}& G3D &FNTU & G3D &FNTU \\
\hline
disorder${1}$&0.1232&0.1273&1.1281&1.1944 \\
%\hline
disorder${2}$&0.1263&0.1330&1.1538&1.2241 \\
%\hline
Our&{\bf 0.1199}&{\bf 0.1210}&{\bf 1.1101}&{\bf 1.1651} \\
\hline
\end{tabular}
\end{table}

\begin{figure}[!t]
\centering
\includegraphics[width=2.5in]{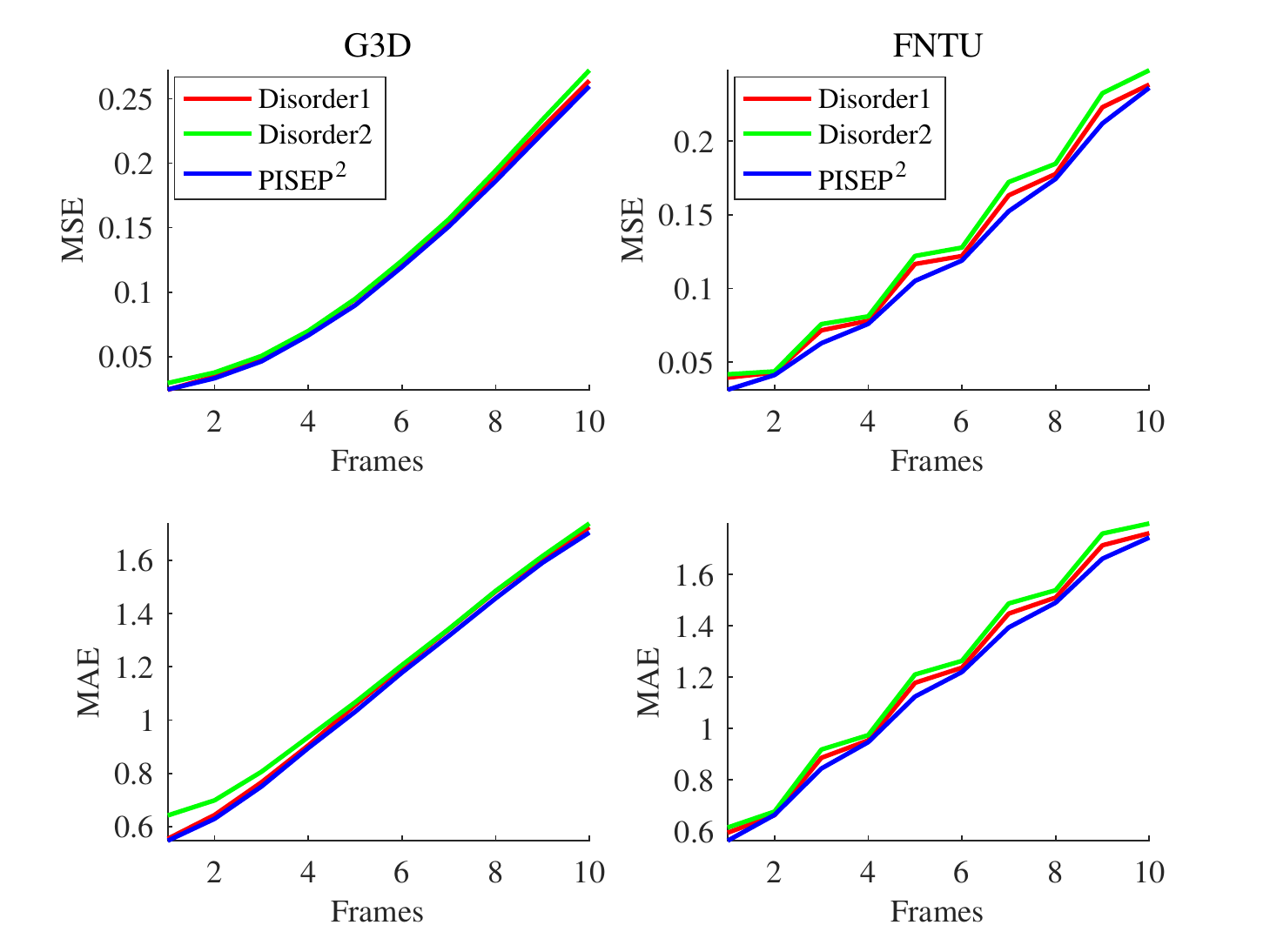}
\caption{Frame-wise performance of different representations.}
\label{fig9}
\end{figure}

Figure \ref{fig9} is the frame-wise performance of different representations. As is shown in Figure \ref{fig9}, the results of our representation are slightly better than the performances of the ``disorder${1}$'' and ``disorder${2}$'' joints representations at all future frames, which demonstrate that the effectiveness of our skeletal representation. But our effectiveness is limited, the possible reason is: due to the small size of our representation, to some extent, all joint can be affected with each other under the convolution operation.

\begin{figure}[!t]
\centering
\includegraphics[width=2.5in]{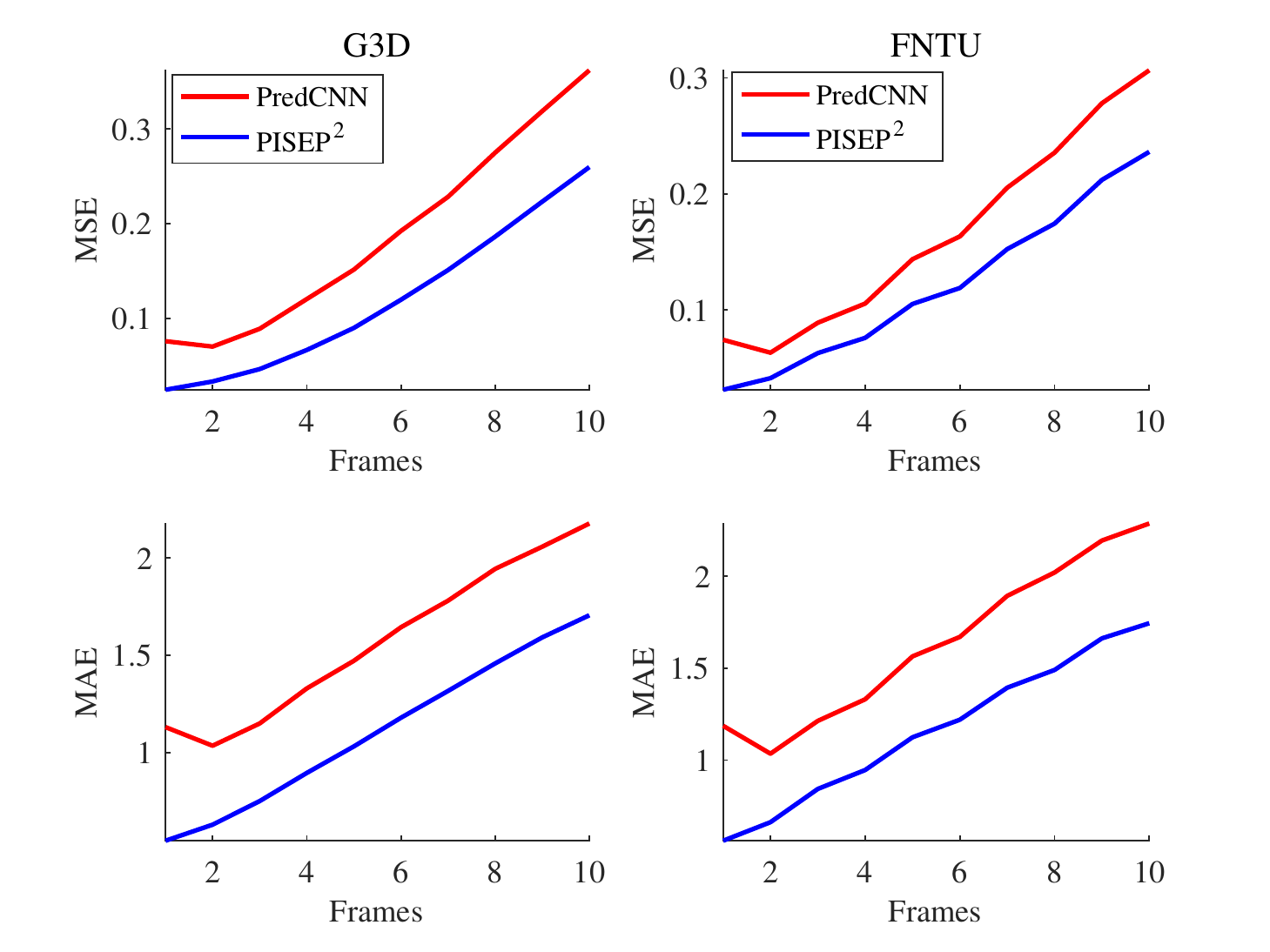}
\caption{Frame-wise performance of different architectures.}
\label{fig10}
\end{figure}

{\bf Evaluation of our network architecture}: PredCNN \cite{predcnn} is the typical representation of the chain network, which is the most similar to ours. To evaluate the effectiveness of our framework, we compare it with PredCNN using our skeletal representation. And the experimental results are shown in Table \ref{table3} and Figure \ref{fig10}. As is shown in Table \ref{table3}, our performance efficiently outperforms PredCNN on both accuracy and computational cost. Our network removed the chain structure proposes to predict all future frames in one step, which can efficiently avoid the error accumulation. The experimental results further evidence the efficiency of our network.

\begin{table}[!t]
\renewcommand{\arraystretch}{1.3}
\caption{Results of different architectures}
\label{table3}
\centering
\begin{tabular}{ccccccc}
\hline
\multirow{2}{*}{Model}& \multicolumn{2}{c}{MSE} & \multicolumn{2}{c}{MAE} & \multicolumn{2}{c}{Test time/ms} \\
 \cline{2-5}& G3D &FNTU & G3D &FNTU & G3D &FNTU \\
\hline
PredCNN\cite{predcnn}&0.1876&0.1665&1.5539&1.6394&4.3828&3.2649 \\
%\hline
PISEP${^2}$&{\bf 0.1199}&{\bf 0.1210}&{\bf 1.1101}&{\bf 1.1651}&{\bf 2.0372}&{\bf 3.1961} \\
\hline
\end{tabular}
\end{table}

Figure \ref{fig10} is the frame-wise performance of different architectures. As is shown in Figure \ref{fig10}, the performance of our network significantly exceeds the performance of PredCNN at all time-steps, especially for the long-term prediction. Because PredCNN is a network with a recursive structure, the output of the current step is the input of the next step. Therefore, the performance of each step is vulnerable to the performance of the previous step. This may lead to the poor performance of PredCNN, especially for the later time-step. The experimental results show that our network can effectively avoid error accumulation again.

{\bf Evaluation of Loss function}: when two poses are too similar, their difference is very small (far less than one). Their square value is smaller. So, in this case, the L${2}$ loss can't guide training well, but the L${1}$ norm loss directly reflects the difference between these two similar poses, which can better guide training. Therefore, we assume that the L${1}$ norm loss better than L${2}$ norm loss for more accurate pose prediction.

\begin{table}[!t]
\renewcommand{\arraystretch}{1.3}
\caption{Performance of different losses}
\label{table4}
\centering
\begin{tabular}{ccccc}
\hline
\multirow{2}{*}{Model}& \multicolumn{2}{c}{MSE} & \multicolumn{2}{c}{MAE} \\
 \cline{2-5}& G3D &FNTU & G3D &FNTU  \\
\hline
PISEP${^2}$ (L${2}$)&0.1434&0.1354&1.4038&1.4063 \\
%\hline
PISEP${^2}$ (L${1}$ norm)&{\bf 0.1199}&{\bf 0.1210}&{\bf 1.1101}&{\bf 1.1651} \\
\hline
\end{tabular}
\end{table}

\begin{figure}[!t]
\centering
\includegraphics[width=2.5in]{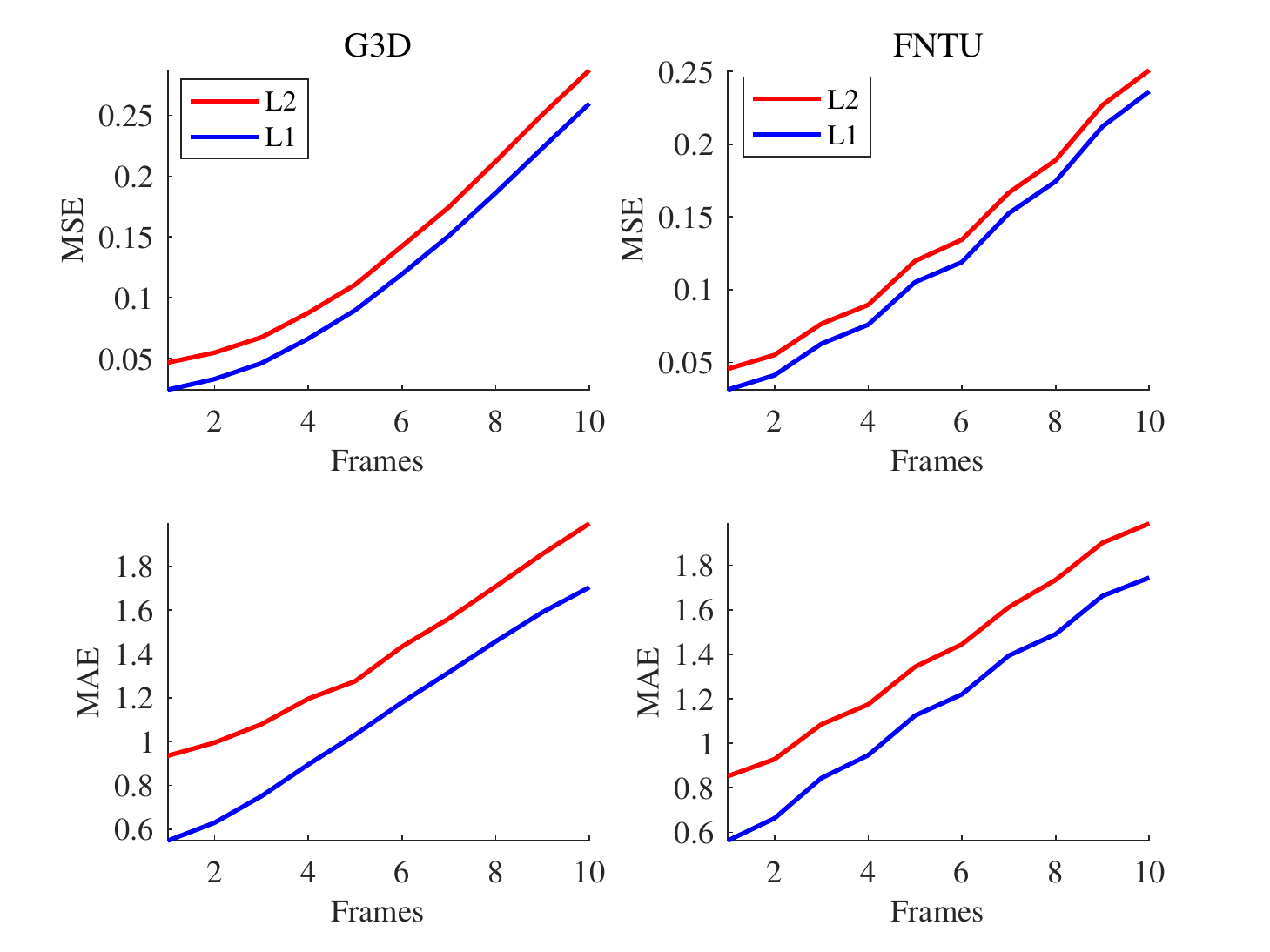}
\caption{Frame-wise performance of different losses.}
\label{fig11}
\end{figure}

To verify the effectiveness of L${1}$ norm loss, we carry out two experiments using our skeletal representation: (${1}$) PISEP${^2}$ with L${2}$ loss; (${2}$) PISEP${^2}$ with L${1}$ norm loss. Table \ref{table4} shows the performance of PISEP${^2}$ with different losses on two challenge datasets. The performance of PISEP${^2}$ with L1 norm loss significantly outperforms it with L${2}$ loss on both datasets. For example, the MSE decreases from ${0.1434}$ to ${0.1199}$, and the MAE decreases from ${1.4038}$ to ${1.1101}$ on G${3}$D. Figure \ref{fig11} is the frame-wise performance of different losses. Compared with L${2}$ loss, the performance of L${1}$ norm loss outperforms of L${2}$ loss at all timestamps on these two datasets. The experimental results further demonstrate the effectiveness of L${1}$ norm loss which is more suitable for more accurate to predict poses.

\subsection{Evaluation of generalization}
The actions on G${3}$D are the gaming related actions, which are different from the actions on FNTU, because most of the actions on FNTU are the daily activity actions. Therefore, to further analyze the performance of our model on unseen data, we present to carry out two groups experiments: (${1}$) pre-trained on FNTU, test on G${3}$D directly; (${2}$) pre-trained on FNTU, fine-tuned on G${3}$D.

The first group experimental results are shown in Table \ref{table5}. And our model shows the best performance on unseen data. Compared with PredCNN, our model decreases by ${0.0986}$ and ${0.8993}$ for MSE and MAE respectively. Compared with S-TE\cite{ButepageDRL}, the MSE and MAE of our model decrease by ${0.0606}$ and ${0.4194}$ respectively. The experimental results show that our model is more general, which is more robust to unseen data. The possible reasons are two folds: (${1}$) {\bf model spatial and temporal information differently}: more precisely, we can model the spatial information using, an LSTM like block, RMBs \cite{vpn} powerfully. And the global temporal information is modeled hierarchically using CMUs \cite{predcnn}. (${2}$) {\bf Non-recursively structure}: different from the commonly used recursive network \cite{predcnn}, we propose to predict all future frames in one step, which can avoid error accumulation efficiently and also improve their predictive performance.

\begin{table}[!t]
\renewcommand{\arraystretch}{1.3}
\caption{Results of general prediction}
\label{table5}
\centering
\begin{tabular}{ccc}
\hline
Model&MSE&MAE \\
\hline
PredCNN\cite{predcnn}&0.2432&2.1706 \\
%\hline
S-TE\cite{ButepageDRL}&0.2052&1.6907 \\
%\hline
PISEP${^2}$ &{\bf 0.1446}&{\bf 1.2713} \\
\hline
\end{tabular}
\end{table}

To further verify the predictive power of our network, we present to carry out the second group experiment. And the experimental results are shown in Table \ref{table6}. Our network achieves optimal experimental results, which is consistent with the experimental results as discussed above.

\begin{table}[!t]
\renewcommand{\arraystretch}{1.3}
\caption{Results of fine-tuned prediction}
\label{table6}
\centering
\begin{tabular}{ccc}
\hline
Model&MSE&MAE \\
\hline
PredCNN\cite{predcnn}&0.1315&1.2808 \\
%\hline
S-TE\cite{ButepageDRL}&0.1289&1.1150 \\
%\hline
PISEP${^2}$ &{\bf 0.1040}&{\bf 0.9379} \\
\hline
\end{tabular}
\end{table}

\begin{figure}[!t]
\centering
\includegraphics[width=2.5in]{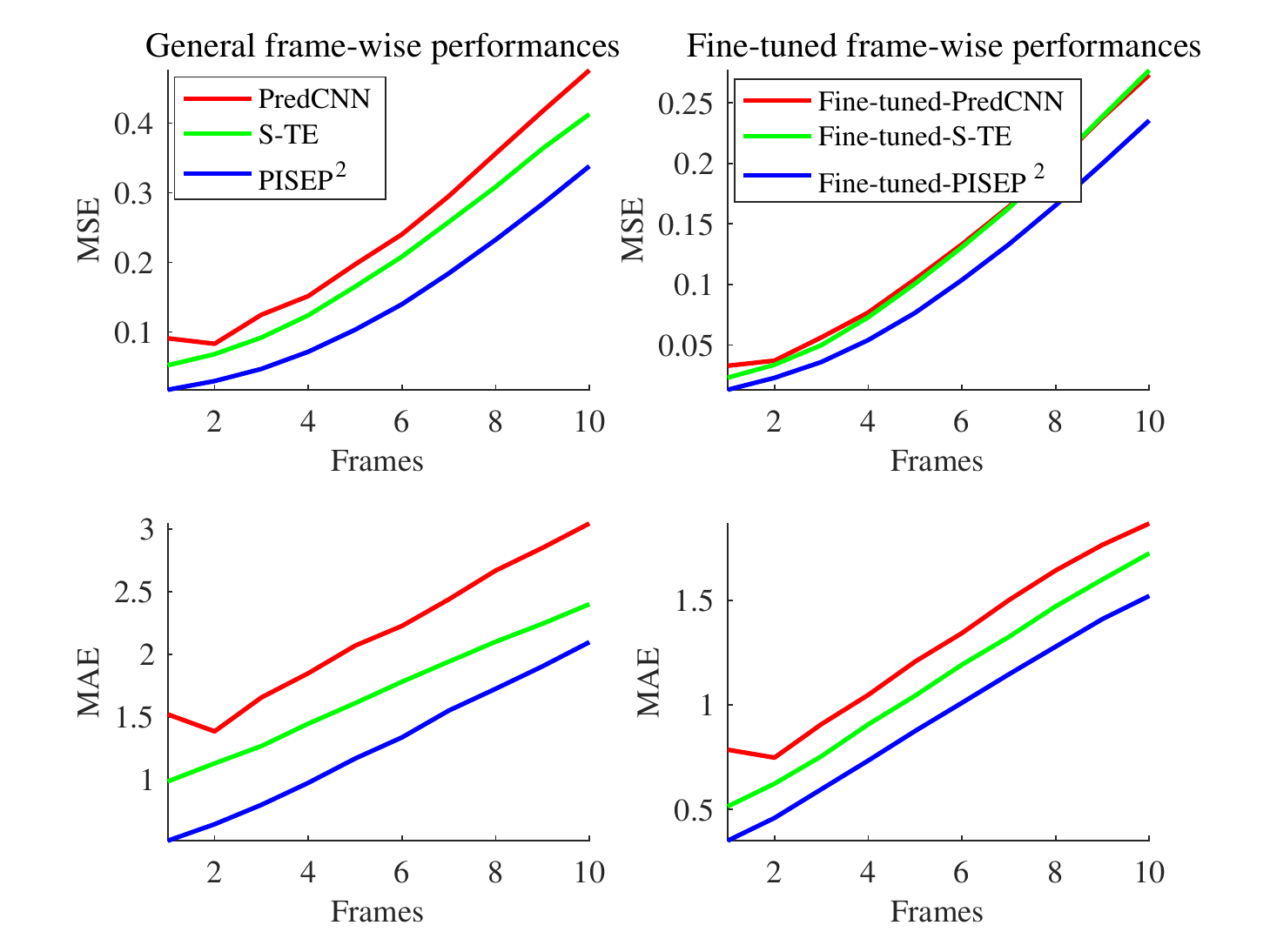}
\caption{Frame-wise performance on unseen data.}
\label{fig12}
\end{figure}

Figure \ref{fig12} is the frame-wise performance on unseen data. Before fine-tuning, as shown in the left part of  Figure \ref{fig12}, our method significantly outperforms all the baselines, which indicates the power of our network to learn the general motion representation. After fine-tuning, as shown in the right part of Figure \ref{fig12}, all of the methods can learn the motion representation on new data well. Moreover, the performance of PredCNN is approximate to the performance of the S-TE. However, our method still surpasses PredCNN and S-TE, especially for the long-term motion prediction, which further evinces the power of our network to efficiently model the temporal evolution of pose sequence.

\begin{figure}[!t]
\centering
\subfloat[]{\includegraphics[width=3.6in]{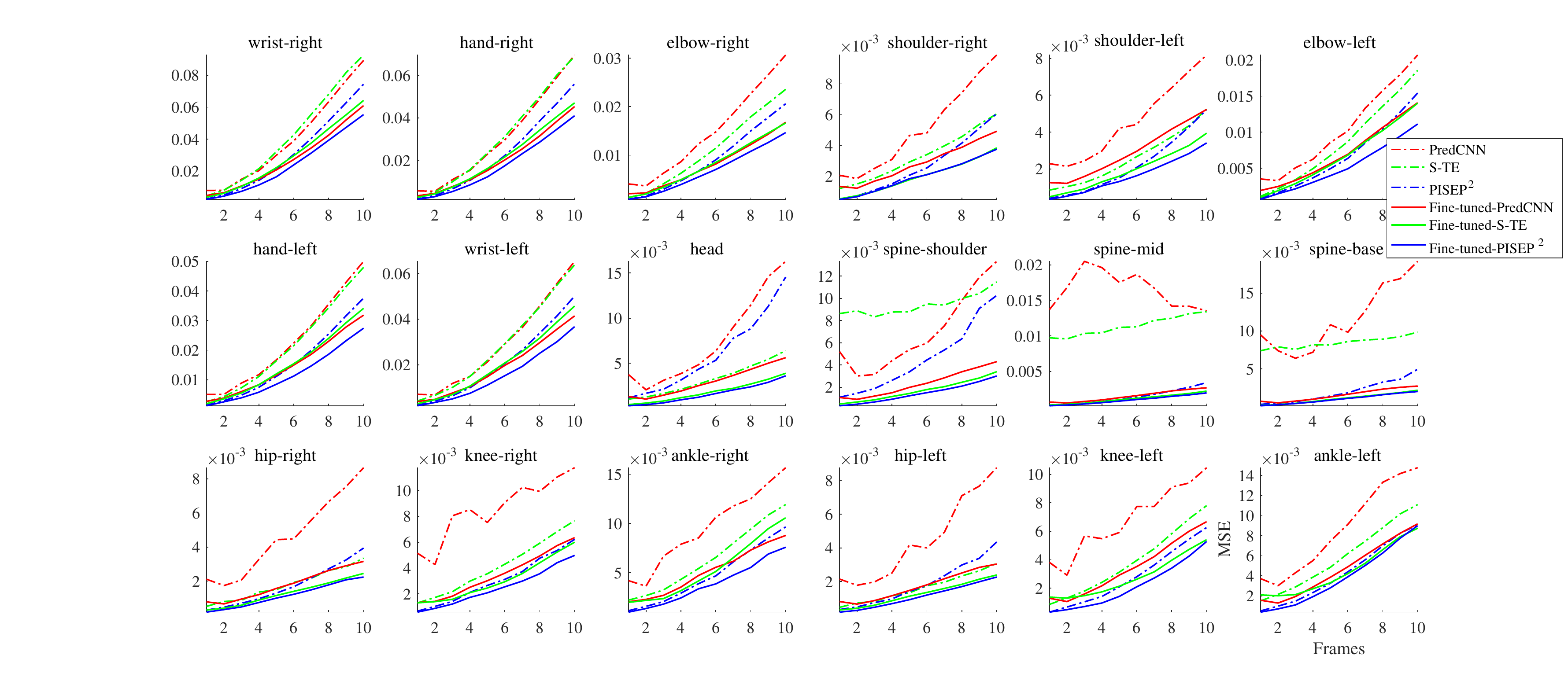}
\label{fig13_1}}
\hfil
\subfloat[]{\includegraphics[width=3.6in]{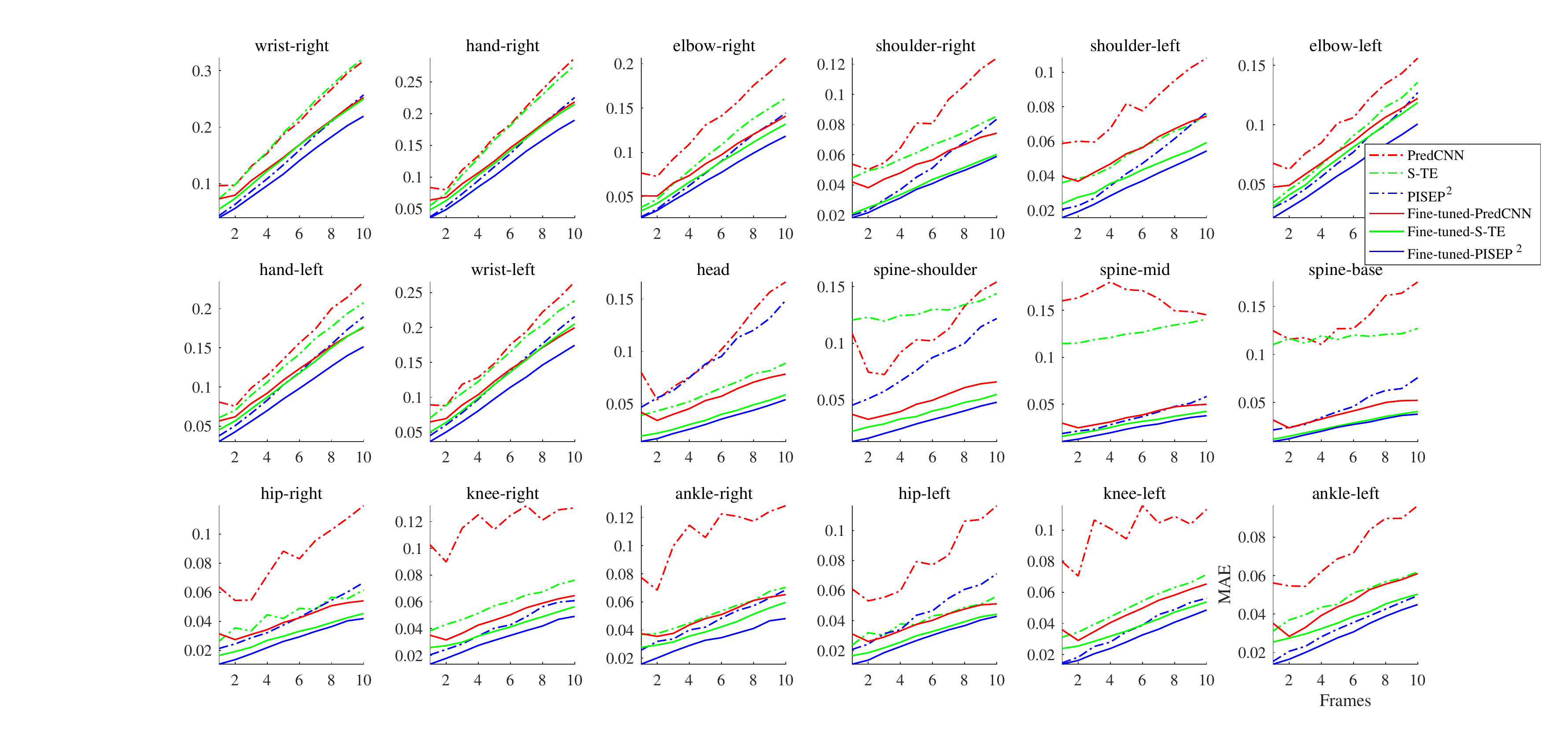}
\label{fig13_2}}
\caption{ Joint-wise performance of unseen data. (a) Joint-wise MSE of general results. (b) Joint-wise MAE of general results. }
\label{fig13}
\end{figure}

To analyze the general performance of our proposed method carefully, joint-wise evaluation is carried out as shown in Figure \ref{fig13}. Similarly, the errors of the joints of the upper limbs are larger than the errors of the joints of the trunk or lower limbs both on MSE and MAE. The possible reason is the same as discussed above. (${1}$) {\bf Before fine-tuning}, the performance of our method better than most of the joints of all the baselines for both MSE and MAE. Among them, the performance of PredCNN achieves the worst performance and is severely unstable. For example, the error of the joint ``spine mid'' fluctuates greatly, and its error at short-term motion prediction is larger than the long-term motion prediction, which is converse to the normal trend. The possible reason is that their recursive structure cause error accumulation, which leads to the poor generalization ability of PredCNN. Compared with PredCNN, the performance of S-TE seems more stable. Because S-TE treats the spatial and temporal information equally, which may not capture the temporal evolution of the pose sequence well. But our model removes the recursive structure, and presents to predict all future poses at one time, which can effectively improve the computation efficiency and avoid error accumulation. Besides, our model significantly outperforms PredCNN and S-TE, which, to a great extent, shows the powerful generalization ability of our network. (${2}$) {\bf After fine-tuning}, all models can learn the specific representation of new data, and our model gains the best performance. This may benefit from our non-chain framework, which can capture the temporal information well and avoid error accumulation.

\subsection{Qualitative Analysis of the Experimental Results}
\begin{figure}[!t]
\centering
\subfloat[]{\includegraphics[width=3.8in]{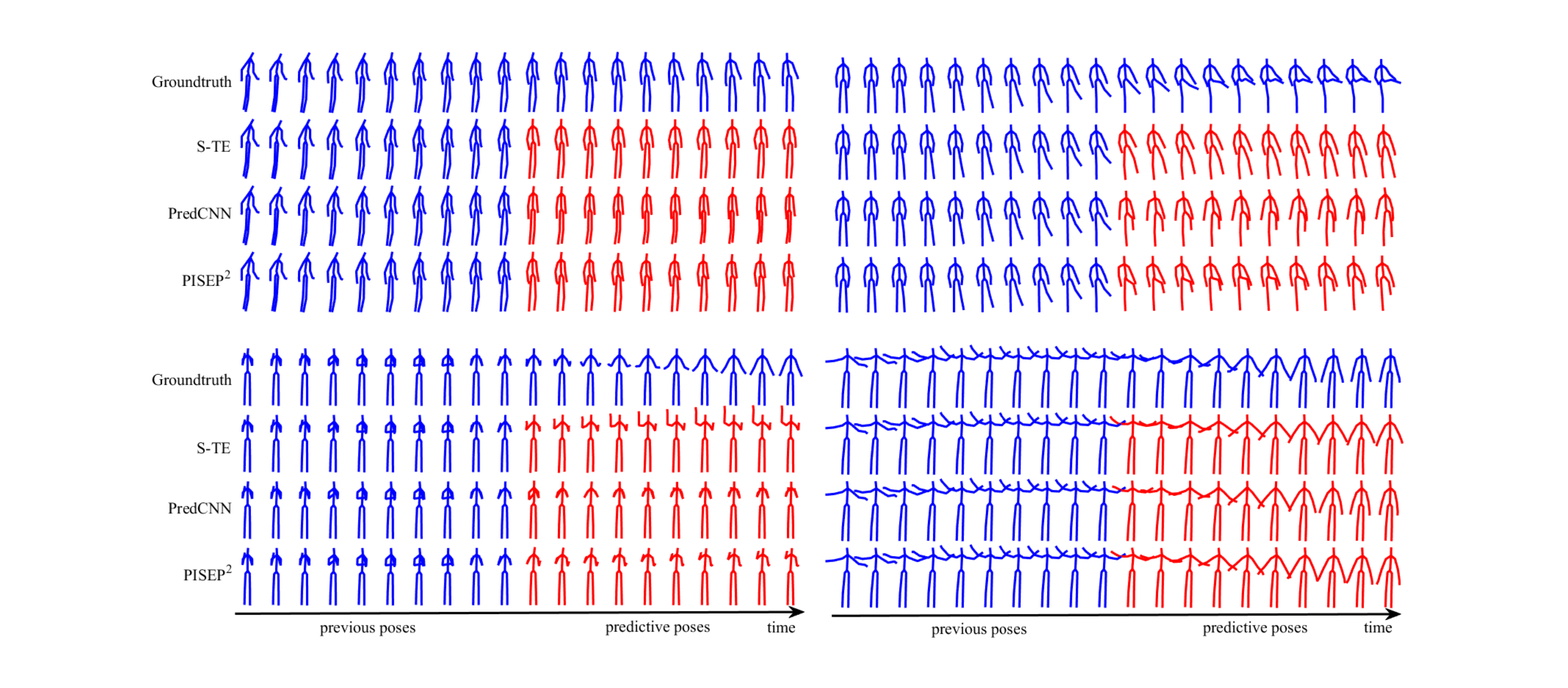}
\label{fig14_1}}
\hfil
\subfloat[]{\includegraphics[width=3.8in]{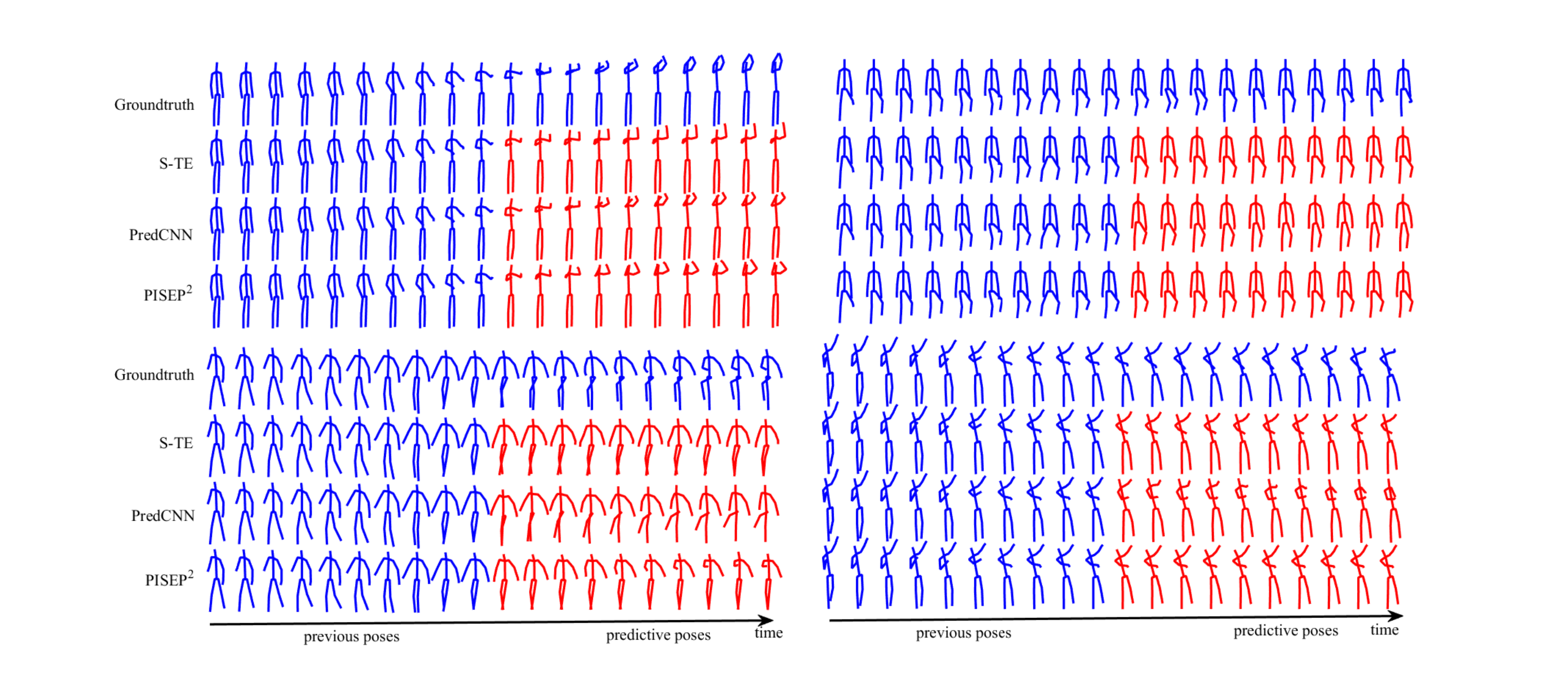}
\label{fig14_2}}
\caption{ Visualization of frame-wise predictive performance. (a) Visualization of frame-wise performance on G${3}$D. (b) Visualization of frame-wise performance on FNTU. }
\label{fig14}
\end{figure}

To show the performance of our proposed method, we visualize the predictive pose frame by frame qualitatively. Figure \ref{fig14} is the visualization of frame-wise performance on two challenging datasets. Here, for each group pose sequences, the first sequence denotes the groundtruth sequence, the second sequence corresponds to the performance of S-TE, the third sequence corresponds to the results of the PredCNN, and the last sequence produces the results of our model. Moreover, all the predictive future poses are marked in red.

As shown in Figure \ref{fig14}, (a) {\bf on G${3}$D}, our model achieves more reasonable performance in general, which further evidences the effectiveness of our proposed method. For example, the top left group pose sequences, the long-term performance of the third sequence performs seems terrible, which is very different from the groundtruth poses. Compared with PredCNN, our predict poses seems more reasonable. For the top right group pose sequences, the S-TE model shows the worst results. The possible reason is that S-TE treats spatial and temporal information equally, which is likely to fail to carefully capture the temporal evolution of the pose sequence. Compared with PredCNN, owe to the non-recurrent structure, our model achieves superior performance, which can avoid error accumulation. Although our result is still far from the groundtruth in the top right and the bottom left sequence, the evolution direction of the pose movement is approximately correct. (b) {\bf On FNTU}, the visualization performance of our method outperforms PredCNN's in general, which demonstrates that our method can efficiently avoid error accumulation. Compared with S-TE, our performance seems more reasonable. For example, for the top left group sequences, although our performance is not good enough, our result seems more reasonable than S-TE. Since S-TE may not distinguish the spatial and temporal information, while our method model the spatial and temporal information with different kinds of blocks, our model can better capture the temporal information than S-TE. This may be the possible reason that our model can predict more reasonable poses.

\begin{figure}[!t]
\centering
\includegraphics[width=3.8in]{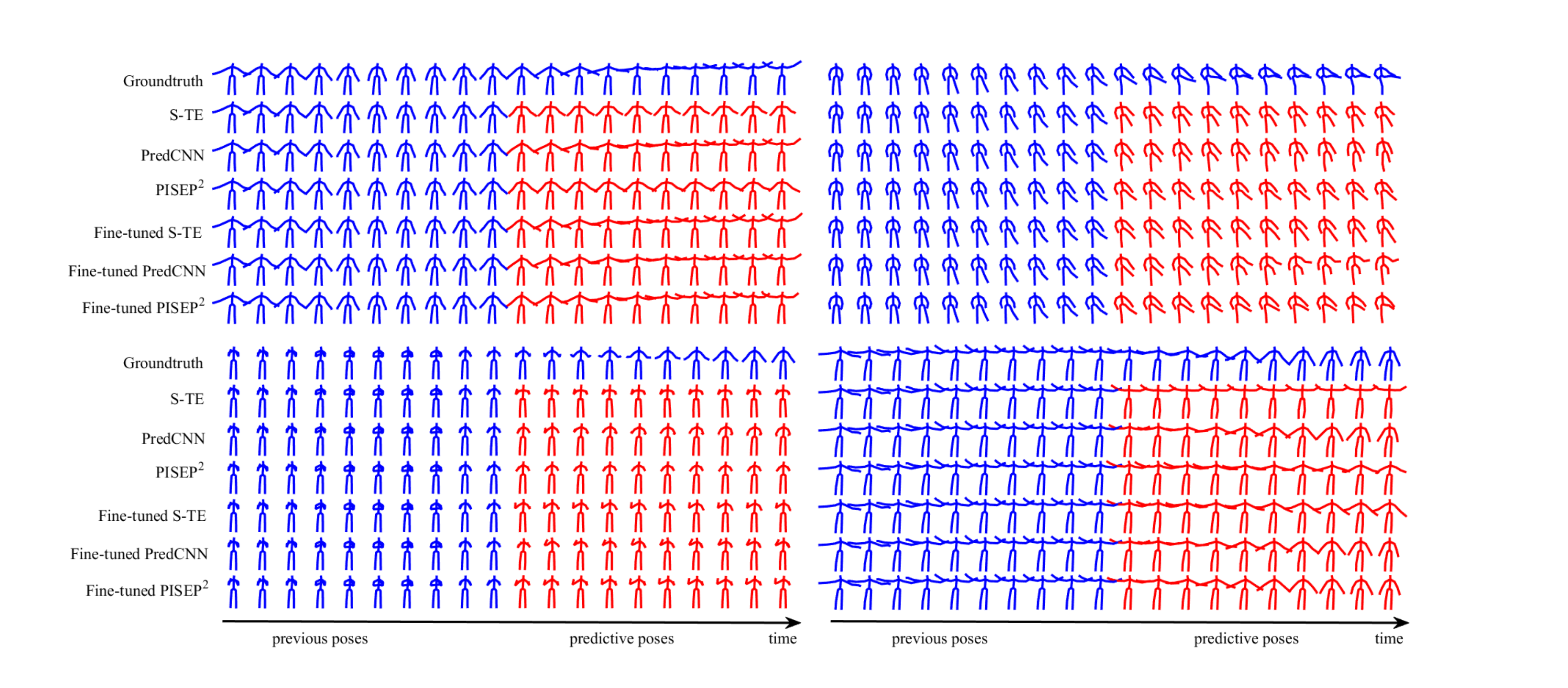}
\caption{Visualization of frame-wise performance on unseen data.}
\label{fig15}
\end{figure}

To further analysis the visualization performance of our model on unseen data, the predictive poses with different models are visualized as shown in Figure \ref{fig15}. (${1}$) {\bf Before fine-tuning}, compared with PredCNN, in most cases, our method achieves better visualization performance. For example, for the top right group pose sequence, our predictive poses are more reasonable since the direction of the motion is almost consistent with the groundtruth. But there are exceptions, such as bottom right group sequences. Our method is slightly worse. Compared with S-TE, our approach is much better. The experimental results show that the generalization performance of our method is much better since our method models spatial and temporal information differently, and can avoid error accumulation efficiently. (${2}$) {\bf After fine-tuning}, the visualization performances of all methods have been improved significantly, but the effect of our approach is the best, which further verify the effectiveness of the proposed method from the side. Among them, there may be some unreasonable phenomena in PredCNN's predictive pose due to the accumulation of errors. For example, for the top right group sequences, the joints of the poses in the long-term prediction seem unreasonable since the evolution direction of the joints is inconsistent with the groundtruth. But our performance is approximately consistent with the groundtruth. This shows again that our method can predict future poses more reasonably.

The best performance of our method, to a great extent, reveals two facts: (${1}$) modeling the spatial and temporal information differently can explore the temporal evolution information better; (${2}$) remove the recursive structure of the sequence to sequence model may can avoid error accumulation efficiently and improve the performance both at accuracy and computation efficiency significantly.



%\subsection{Subsection Heading Here}
%Subsection text here.

\section{Conclusion}
In this work, we have formulated a new problem of ${3}$D pose prediction using joint coordinate sequence data and proposed a conceptually simple but efficient framework to address this new problem. Specifically, we provide a conveniently ${3}$D pose prediction method, which can efficiently evaluate and visualize its performance. Furthermore, we present a new skeletal representation, which can conveniently model diverse correlations of different limbs, the local characteristic of the human body, and the global temporal evolution of previous poses. Besides, a new sequence to sequence model is proposed to predict all future frames in one step, and also achieves the state-of-the-art performance, which can significantly improve the computational efficiency and avoid error accumulation. In sum, we have shown the effectiveness of the proposed new skeletal representation and the proposed framework, which can provide an efficient pose prediction method.

\section*{Acknowledgment}

This work was supported partly by the National Natural Science Foundation of China (Grant No. 61673192), the Fund for Outstanding Youth of Shandong Provincial High School (ZR2016JL023), the Basic Scientific Research Project of Beijing University of Posts and Telecommunications (2018RC31), and BUPT Excellent Ph.D. Students Foundation (CX2019111). The research in this paper used the NTU RGB+D Action Recognition Dataset made available by the ROSE Lab at the Nanyang Technological University, Singapore.

\ifCLASSOPTIONcaptionsoff
  \newpage
\fi



\bibliographystyle{IEEEtran}
\bibliography{IEEEabrv,PISEP2}
\end{document}